\documentclass[12pt]{article}
\usepackage{amssymb,amsmath,amsfonts,eurosym,geometry,ulem,color,setspace,sectsty,comment,footmisc,natbib,pdflscape,array}
\usepackage{booktabs}
\usepackage{multirow}
\usepackage{soul}
\usepackage{xcolor}
\usepackage{siunitx}
\usepackage{csvsimple}
\usepackage{ragged2e}
\usepackage{caption}
\usepackage{subcaption}
\usepackage{graphicx}
\usepackage{placeins}
\usepackage{tikz}
\usepackage{hyperref}
\usetikzlibrary{arrows.meta, positioning, calc, fit}
\graphicspath{{figures/}}
\DeclareGraphicsExtensions{.pdf,.png,.jpg}
\newcolumntype{L}[1]{>{\raggedright\let\newline\\arraybackslash\hspace{0pt}}m{#1}}
\newcolumntype{C}[1]{>{\centering\let\newline\\arraybackslash\hspace{0pt}}m{#1}}
\newcolumntype{R}[1]{>{\raggedleft\let\newline\\arraybackslash\hspace{0pt}}m{#1}}
\begin{document}

\begin{titlepage}
\title{Interpretable Machine Learning for Predicting Startup Funding, Patenting, and Exits}
\author{
Saeid Mashhadi\thanks{LeBow College of Business, Drexel University, Philadelphia, PA 19104 (email: \href{mailto:saeid.mashhadi@drexel.edu}{saeid.mashhadi@drexel.edu})} 
\and Amirhossein Saghezchi\thanks{LeBow College of Business, Drexel University, Philadelphia, PA 19104} 
\and Vesal Ghassemzadeh Kashani\thanks{University of Tehran, Tehran, Iran}
}

\date{June 19, 2026}
\maketitle
\begin{abstract}

This study develops an interpretable machine learning framework to forecast startup outcomes, including funding, patenting, and exit. A firm-quarter panel for 2010-2023 is constructed from Crunchbase and matched to U.S. Patent and Trademark Office (USPTO) data. Three horizons are evaluated: next funding within 12 months, patent-stock growth within 24 months, and exit through an initial public offering (IPO) or acquisition within 36 months. Preprocessing is fit on a development window (2010-2019) and applied without change to later cohorts to avoid leakage. Class imbalance is addressed using inverse-prevalence weights and the Synthetic Minority Oversampling Technique for Nominal and Continuous features (SMOTE-NC). Logistic regression and tree ensembles, including Random Forest, XGBoost, LightGBM, and CatBoost, are compared using the area under the precision-recall curve (PR-AUC) and the area under the receiver operating characteristic curve (AUROC). Patent, funding, and exit predictions achieve AUROC values of 0.921, 0.817, and 0.872, providing transparent and reproducible rankings for innovation finance.

\vspace{0.2in}

\bigskip
\end{abstract}
\setcounter{page}{0}
\thispagestyle{empty}
\end{titlepage}
\pagebreak \newpage

\doublespacing


\section{Introduction} \label{sec:introduction}

Innovation is central to value creation in entrepreneurship. For high-growth startups, the ability to generate and protect new knowledge shapes product-market paths, affects access to external finance, and influences the likelihood of timely exit. Patents are central to this process, as they secure appropriability and act as credible signals that reduce information frictions between founders and investors \citep{Conti2013JOIE,HsuZiedonis2013SMJ}. Causal evidence shows that patent grants improve a venture’s prospects by increasing financing access, growth, and the probability of IPO or acquisition \citep{FarreMensa2020JF}. At the ecosystem level, venture capital is associated with innovative output, reinforcing a cycle between invention, investor attention, and firm growth \citep{KortumLerner2000RAND}. The ability to forecast which startups will expand their intellectual property portfolios, raise additional capital, and reach an exit has direct value for investors, corporate acquirers, and policy stakeholders.

Predictive work in entrepreneurial finance and strategy generally follows two strands. One focuses on financing outcomes, such as the timing and likelihood of later rounds, based on observable histories of capital raised, investor breadth, and founder characteristics \citep{Gompers2010JFE,Eggers2015AMJ,Gompers2020JFE}. The other uses innovation as an explanatory factor, showing that patent stocks and citations anticipate financing and exits \citep{HsuZiedonis2013SMJ,FarreMensa2020JF,Kogan2017QJE}.

Recent work uses large entrepreneurial datasets to forecast startup performance with machine learning, showing that ensemble models can recover meaningful nonlinearities in funding and survival \citep{Kim2023TFSC,Arroyo2019IEEE}. Venture capital research also shows that investors often rely on signals that are measurable in digital traces and patent portfolios, and that quantitative models can approximate or complement VC decision processes \citep{LyonnetStern2024WP}. These studies indicate that predictive approaches can help disentangle the relative importance of financing momentum, founder maturity, and innovation intensity for near-term and medium-horizon outcomes. Moreover, recent machine learning (ML) studies use large platforms such as Crunchbase to predict survival, fundraising, or valuation by capturing non-linearities and high-dimensional interactions \citep{Antretter2019JBVI,Dalle2017OECD}. Parallel work in corporate finance shows that interpretable machine learning can recover meaningful drivers while delivering competitive out-of-sample accuracy. For example, \citet{Kim2024} use a model zoo with imputation, oversampling, and SHAP-based explanations to predict activist targets.

Two gaps remain in the literature. First, finance- and innovation-focused predictions are often studied separately. This separation leaves open questions about the joint and relative roles of financing recency and momentum versus intellectual property stocks in shaping near-term funding, medium-horizon patenting, and longer-horizon exit. Second, while ML can improve discrimination, many applications do not enforce leakage-safe time splits or provide transparent interpretability and reliability checks. Scalable and interpretable prediction of startup patenting, defined as whether a firm will expand its patent stock over two years, has received less systematic attention than fundraising or exit forecasting, despite its importance.

\paragraph{This paper.} An interpretable ML pipeline is developed to forecast three outcomes on a leakage-safe firm-quarter panel: (i) next financing within 12 months, (ii) patent-stock growth within 24 months, and (iii) IPO or acquisition within 36 months. The panel was built from Crunchbase and merged to U.S. Patent and Trademark Office (USPTO) assignee data to track cumulative patent and citation stocks \citep{Dalle2017OECD,Kogan2017QJE}. All steps were scripted in Python for exact reproducibility \citep{McKinney2010}. To avoid look-ahead, preprocessing (infinite-to-NaN, missing-value flagging, and median imputation) was fit on a development window (2010-2019) and applied without change to a holdout window (2020-2021) and a final window (2022-2023). Class imbalance was addressed inside development with inverse-prevalence weights and Synthetic Minority Oversampling Technique for Nominal and Continuous features (SMOTE-NC), and evaluation splits were not resampled or reweighted. A linear baseline (regularized logistic) was compared with tree ensembles (Random Forest \citep{Breiman2001}, XGBoost, LightGBM, CatBoost). Winners were selected by area under the precision-recall curve (PR-AUC) with area under the receiver operating characteristic curve (AUROC) as a tiebreaker. Models were diagnosed with interpretable artifacts: Tree-based SHapley Additive exPlanations (TreeSHAP) for gradient-boosted trees where feasible and permutation or impurity importance for Random Forests, along with calibration curves and partial dependence \citep{LundbergLee2017NeurIPS}.

The unit of analysis is the firm-quarter. Outcomes were defined strictly forward from each quarter-end, and rows too close to panel boundaries for a full horizon were flagged as non-evaluable. The time splits are non-overlapping: 2010-2019 (development), 2020-2021 (holdout), and 2022-2023 (final). Predictor construction relied on signals highlighted in prior work on startup dynamics, including financing recency and momentum, cumulative exposure to investors and capital, and intellectual property stocks and citations \citep{Gompers2010JFE,Conti2013JOIE,HsuZiedonis2013SMJ,FarreMensa2020JF}. The development feature list is persisted to guarantee column alignment at scoring time, and deterministic seeds, filenames, and manifests are used to enable replication.

Three results summarize the contribution. First, \emph{startup patenting is highly predictable at the firm-quarter level}. On the holdout window, the Random Forest trained with inverse-prevalence weights achieves AUROC $=0.921$ and PR-AUC $=0.631$ for predicting a two-year increase in the patent stock. Global importance and partial-dependence diagnostics show pronounced mean reversion with respect to existing IP stock and a negative gradient with time since last round, which is consistent with the view that younger, actively financed firms are more likely to expand their portfolios.

Second, \emph{fundraising within 12 months is dominated by recency and momentum}. On the final window, a LightGBM model with inverse-prevalence weights attains AUROC $=0.817$ and PR-AUC $=0.220$. SHAP identifies days since last round and age as first-order drivers, with concave gains from cumulative capital and investor breadth. Reliability curves show optimism at higher scores. Isotonic recalibration is therefore recommended when calibrated probabilities, rather than rankings, are required.

Third, \emph{exits are associated with financing maturity}. For 36-month exits, the Random Forest with inverse-prevalence weights delivers AUROC $=0.872$ and PR-AUC $=0.559$ on holdout. Importance and partial dependence indicate that time since last round and age, together with cumulative capital raised, are the leading drivers of exits, with investor breadth and cumulative rounds contributing secondarily. These patterns are consistent with evidence that VCs behave as selective experimenters. High-growth outcomes are rare but exhibit strong dependence on accumulated capital, market validation, and early technological signals \citep{NandaRhodesKropf2013JFE}. Studies that compare machine learning predictions to actual VC investment choices also find that VCs often fund firms that rank highly in predictive models, although some underserved high-potential firms persist (Lyonnet and Stern, 2024). This alignment supports the external validity of the predictors identified here, particularly financing maturity and recent fundraising intensity. Together, these results show that interpretable ML can deliver ranking-useful precision on out-of-time cohorts while recovering mechanisms that align with economic priors. Fundraising behaves as a recency and momentum phenomenon, patent growth reflects maturity and mean reversion, and exits load on financing depth.

\paragraph{Relation to the literature.} The analysis complements and integrates three literatures. From entrepreneurial finance, roles for founder and firm maturity and for financing histories in shaping outcomes are confirmed \citep{Gompers2010JFE,Eggers2015AMJ,Gompers2020JFE}. From the innovation literature, patent stocks and citations are used as predictive inputs rather than only as outcomes, consistent with patents both signaling quality and causally enhancing financing and growth \citep{Conti2013JOIE,HsuZiedonis2013SMJ,FarreMensa2020JF,Kogan2017QJE}. From ML in finance and innovation, an interpretable, leakage-averse pipeline is adopted in the spirit of \citet{Kim2024}, targeted to the startup domain with quarter-level panels, non-overlapping time splits, and explicit reliability checks. The novelty is the combination of a patent-growth dependent variable at scale, unified benchmarking across funding, patenting, and exit horizons within one pipeline, and deployable ranked lists from strictly out-of-time scoring.

\paragraph{Contributions.} The paper makes four contributions:
\begin{itemize}
\item \textbf{An interpretable, leakage-safe pipeline for startup outcomes.} A firm-quarter panel (2010-2023) with strict non-overlapping time splits, development-only preprocessing, and persisted feature lists is constructed, providing a reproducible benchmark that integrates funding, patenting, and exit horizons \citep{Dalle2017OECD,McKinney2010}.
\item \textbf{Scalable prediction of patent-stock growth.} Two-year patent expansion is shown to be predictable out of time (AUROC $=0.921$, PR-AUC $=0.631$), with interpretable drivers that align with economic priors on maturity and financing recency \citep{Kogan2017QJE,FarreMensa2020JF}.
\item \textbf{Unified model benchmarking with transparent diagnostics.} Logistic and modern tree ensembles are compared under imbalance treatments. Selection is by PR-AUC, and SHAP or importance, partial dependence, and calibration curves are reported to pair discrimination with explainability \citep{Breiman2001,LundbergLee2017NeurIPS,Kim2024}.
\item \textbf{Deployable ranked target lists.} Validated models are translated into out-of-time scored cohorts for screening and benchmarking, and the conditions for post hoc calibration are clarified for probability-based decisions.
\end{itemize}

The remainder of this study is structured as follows. Section~\ref{sec:literature} discusses related work. Section~\ref{sec:data} details data collection (Crunchbase-USPTO merge), preprocessing, imbalance handling, and model families. Section~\ref{sec:result} reports the results, including descriptive diagnostics, model selection, interpretability, calibration, and out-of-time scoring manifests. Section~\ref{sec:conclusion} concludes.

\section{Literature Review} \label{sec:literature}

Research on predicting startup outcomes spans finance, entrepreneurship, and innovation, and has shifted from traditional statistical analyses to broader predictive frameworks. Early work identified observable correlates of startup success or failure, including the timing and size of funding rounds, founder experience, and market conditions. Venture investors have consistently emphasized founding-team quality as a main factor in investment decisions \citep{Gompers2020JFE}. Empirical work supports this view, as founders with prior success tend to succeed again, while experience from failure provides a weaker benefit \citep{Gompers2010JFE,Eggers2015AMJ}. Beyond founder experience, recent predictive studies use large entrepreneurial databases to forecast startup outcomes. \citet{Kim2023TFSC} apply machine learning to Crunchbase firms and find that recency of financing, team size, and investor networks dominate survival predictions. \citet{Arroyo2019IEEE} show that ensemble-tree models outperform linear baselines when VCs attempt to screen promising ventures at scale. These results reinforce the idea that high-frequency financial and organizational signals carry predictive content for short-horizon funding and survival dynamics. Complementing predictive studies, \citet{Saghezchi2024JBMS} develop an optimization model for scale-ups that allocates cash flows across investing, operating, and financing activities, combining a Markowitz mean–variance setup with behavioral reinforcement updates to stabilize performance through the transition phase. These results suggest that human capital and entrepreneurial persistence, together with early financing dynamics, are core elements behind heterogeneity in startup performance.

A second stream studies innovation outputs, especially patents and their quality, as predictors of venture success. Patents play a dual role by protecting intellectual property and signaling technological quality, which helps reduce information asymmetries between entrepreneurs and investors \citep{Conti2013JOIE}. Evidence shows that startups holding patents, particularly highly cited ones, are more likely to secure additional financing and reach successful exits \citep{HsuZiedonis2013SMJ,FarreMensa2020JF}. Quasi-experimental evidence provides causal support, as \citet{FarreMensa2020JF} exploit quasi-random examiner assignment and show that favorable patent grants are followed by greater growth, improved access to external capital, and higher chances of IPO or acquisition. These findings indicate that patents do more than correlate with success, as they also shape a venture’s trajectory. At the aggregate level, venture capital investment is associated with higher innovative output, creating a reinforcing cycle between innovation, investor attention, and firm growth \citep{KortumLerner2000RAND}. Overall, innovation indicators appear central for predicting startup outcomes.

Complementing these findings, empirical studies show that the informational content of patents goes beyond counts or citations. \citet{Haeussler2014RP} document that patent applications and grants help VCs assess technological maturity in early-stage ventures. \citet{Xu2024TFSC} find that the economic value of early patents predicts subsequent innovation trajectories. Together, these studies indicate that patent-based measures can forecast both financing and innovation outcomes, which motivates the inclusion of patent-stock dynamics as a dependent variable.


The availability of large-scale datasets such as Crunchbase and PitchBook has encouraged the use of ML to forecast startup outcomes. ML is well suited to this setting because it captures non-linearities and high-dimensional interactions that logit or probit models may miss. Common targets include next-round financing, survival, valuation milestones, and exit events. Evidence indicates that tree-based ensembles perform well on these tasks. \citet{Antretter2019JBVI} use Twitter-based measures of online legitimacy to predict five-year survival and show that social attention adds predictive content beyond standard covariates. Related work in corporate finance shows the value of interpretable ML, as \citet{Kim2024} develop interpretable random-forest models with SHAP-based explanations to predict activist fund targets and show that SHAP recovers economically meaningful drivers such as valuation and free float. Outside finance, interpretable gradient-boosting models such as LightGBM combined with SHAP also demonstrate strong performance in complex nonlinear prediction tasks, as shown by \citet{Sakhaee2025JEnvMan} in forecasting electrochemical oxidation efficiency. Post hoc interpretability tools such as SHAP are used to attribute predictions in high-capacity models, which improves transparency and comparison with theory \citep{LundbergLee2017NeurIPS}. Together, these studies suggest that ML can improve out-of-sample accuracy in high-dimensional settings while preserving interpretable economic insights.

Despite these advances, two limitations persist. Financial and innovation factors have often been studied separately, as innovation-focused work emphasizes patenting outcomes while finance-oriented work focuses on funding dynamics, which leaves open questions about their joint and relative importance. In addition, few studies benchmark modern ML methods against econometric alternatives under strict out-of-sample validation while also requiring interpretability. The present study addresses these gaps by integrating financial and innovation features within a single ML framework, benchmarking ensemble-tree methods against baseline logistic models, and applying SHAP analysis to recover economically meaningful drivers. The goal is to show that interpretable ML can raise predictive accuracy and clarify why innovation signals and financial traction forecast funding, innovation, and exit outcomes. Efforts to integrate interpretability into predictive finance include the use of SHAP-based diagnostics in high-dimensional credit and risk applications \citep{Bussmann2021CE}. These studies show that post hoc attribution tools can preserve economic interpretability while maintaining competitive predictive accuracy. Similar concerns arise in innovation and venture settings, where models must balance flexibility, transparency, and strict respect for temporal ordering to avoid look-ahead bias \citep{Bergmeir2018CSDA, Cerqueira2020ML}.

\section{Research Methodology} \label{sec:data}
\subsection{Data Collection}
A U.S. startup financing panel was constructed from Crunchbase covering $222{,}126$ funding rounds announced between 2010 and 2023.\footnote{Crunchbase is a widely used, continuously updated directory of entrepreneurial activity. See, e.g., \citet{Dalle2017OECD} for documentation and validation of its research use.} The unit of observation is the \emph{funding round}. Each record is identified by a unique funding-round identifier linked to an organization identifier and a standardized issuer name. Rounds with missing organization identifiers were dropped, and duplicate rows were removed by exact match on the funding-round identifier. For each round, the announcement date, investment type (e.g., seed, venture, grant, debt), and stage category (early-stage, mid-stage, late-stage, other) were observed. Investor participation was tracked using the number of investors per round and the set of lead investors when available. Exit outcomes were proxied by platform flags for initial public offerings (IPO) and acquisitions. The round-level data are transformed into a firm-quarter panel in \S\ref{sec:data-processing}.

To capture innovative activity, the Crunchbase panel was merged with USPTO data. Organizations were linked to their granted patents using assignee information and standard name-matching procedures, which allowed observation of both patent counts and forward citations at the startup level. These variables serve as proxies for the quantity and quality of technological output, following prior work on patent-based measures of innovation \citep{Kogan2017QJE}.

The dataset yielded a median round size of \$1.76 million (mean \$20.99 million; 95th percentile \(\approx\)\$63.5 million) and a median of one participating investor. California, Massachusetts, and New York account for the largest shares of issuers, while the most frequent industry labels are \emph{Consumer goods and services}, \emph{Computer software and Internet}, and \emph{Biotechnology and health care}. Exit flags indicate that 6.64\% of issuer-round pairs are associated with an IPO and 14.91\% with an acquisition in the broader firm history. These patterns align with stylized facts of entrepreneurial finance, including skewed capital distributions, geographic concentration in a few hubs, and sparse binary exit outcomes.


\subsection{Data Processing}\label{sec:data-processing}

\begin{figure}[!htbp]
\centering

\resizebox{\textwidth}{!}{
\begin{tikzpicture}[
  node distance = 4mm and 6mm,
  >=Latex,
  every node/.style = {font=\footnotesize},
  box/.style={draw, rounded corners=6pt, thick, fill=blue!15,
            align=left, inner sep=6pt, text width=0.19\textwidth,
            minimum height=45mm},
  ribbon/.style= {draw, rounded corners=6pt, thick, fill=blue!25,
                  align=left, inner sep=6pt, text width=0.95\textwidth}
]

\node[box] (prep)       {\textbf{1. Data Preparation \& Imputation}\\
-- Standardize IDs \& timestamps\\
-- Replace $\infty$ with NaN\\
-- Median imputation (dev only)\\
-- Persist feature list};
\node[box, right=of prep] (imbalance) {\textbf{2. Class Imbalance Handling}\\
-- Inverse-prob.\ weights (dev only)\\
-- Random oversampling (ROS)\\
-- SMOTE-NC / Borderline-SMOTE / ADASYN};
\node[box, right=of imbalance] (modeldev) {\textbf{3. Model Development \& Selection}\\
-- Logistic, Random Forest\\
-- XGBoost / LightGBM / CatBoost\\
-- Select by PR-AUC (tie: AUROC)};
\node[box, right=of modeldev] (analysis) {\textbf{4. Model Analysis}\\
-- TreeSHAP importances\\
-- Calibration (curves, Brier)\\
-- Partial dependence (top features)};
\node[box, right=of analysis] (prediction) {\textbf{5. Out-of-Sample Prediction}\\
-- Forward cohorts ($h=12/24/36$ mo)\\
-- Align columns via feature list\\
-- Apply dev medians; output $\hat{p}$};

\draw[->, thick, shorten >=2pt, shorten <=2pt] (prep.east) -- (imbalance.west);
\draw[->, thick, shorten >=2pt, shorten <=2pt] (imbalance.east) -- (modeldev.west);
\draw[->, thick, shorten >=2pt, shorten <=2pt] (modeldev.east) -- (analysis.west);
\draw[->, thick, shorten >=2pt, shorten <=2pt] (analysis.east) -- (prediction.west);

\node[inner sep=0pt, fit=(prep)(prediction)] (rowfit) {};
\node[ribbon, below=12mm of rowfit.south] (rib)
{\textbf{Leakage Controls \& Reproducibility}\\
-- Preprocessing estimated on development only; reused unchanged on holdout/final\\
-- Resampling confined to development; none on evaluation data\\
-- Deterministic filenames; saved leaderboards, figures, models};


\end{tikzpicture}
}

\caption{End-to-end data processing and modeling pipeline. Steps~1–5 summarize the sequential workflow from data preparation to out-of-sample prediction. The lower ribbon highlights leakage controls and reproducibility safeguards, including development-only preprocessing, resampling confined to training data, and deterministic artifacts for replication.}
\label{fig:pipeline}
\end{figure}
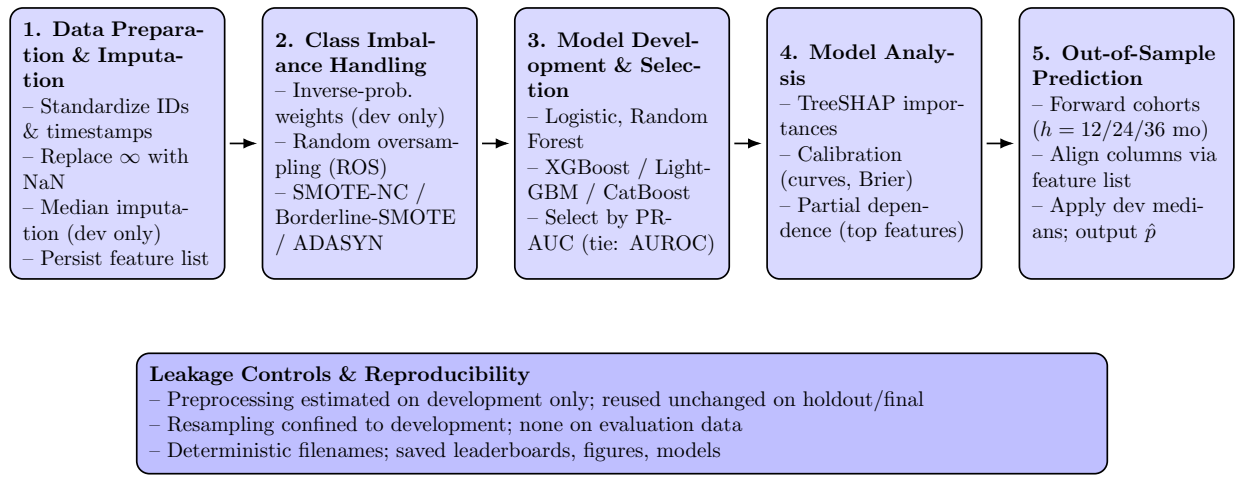

Figure~\ref{fig:pipeline} (Box~1) provides an overview of the transformation from raw Crunchbase extracts to a leakage-safe firm--quarter panel for 2010--2023. All steps were scripted in Python and versioned for exact reproducibility \citep{McKinney2010}. Platform identifiers (\texttt{org\_uuid}) and quarter-end dates were retained to preserve a lossless lineage \citep{Dalle2017OECD}.

The panel was created by building a quarterly calendar for each firm, beginning at the first observed activity (or the founding date, when available) and ending at the earliest exit or December~31,~2023. Periods following an IPO or acquisition were excluded to prevent post-outcome contamination. Financing activity was aggregated to the quarter level, producing counts of rounds, participating investors, and total capital raised. To reduce heterogeneity in platform labels, investment types were classified into four categories: early-stage, mid-stage, late-stage, and other. Cumulative histories were constructed for total capital raised, number of rounds, and number of investors, together with momentum measures (e.g., funding raised in the prior four quarters) and recency measures (days since the last round).

Innovation variables were merged from matched USPTO data. Patent counts and citation totals were aligned to the firm--quarter calendar and \emph{carried forward within firm histories} (with no backfilling prior to a firm’s first observation), creating cumulative stocks that serve as proxies for technological output \citep{Kogan2017QJE}. Firm descriptors such as industry, geography, and founding date were attached, and firm age in years was computed.

Outcomes were defined strictly forward from the quarter boundary. Three dependent variables were created: (i) whether the firm raises another funding round within 12 months, (ii) whether its patent stock increases within 24 months, and (iii) whether it experiences an IPO or acquisition within 36 months. Observations that are too close to the panel boundary for a full horizon were marked as non-evaluable and excluded from the corresponding outcome. Temporal splits follow Figure~\ref{fig:pipeline}: development (2010--2019), holdout (2020--2021), and final (2022--2023). All preprocessing parameters in Box~1 were estimated on the development period only and then applied forward without change.

Missingness was handled within Box~1 by replacing infinities with \texttt{NaN}, adding explicit indicators for features with $\geq$10\% NA in development \emph{plus a forced flag for days since last round}, and applying median imputation fitted on development and reused unchanged on holdout and final. To ensure column alignment across splits, the development feature list was persisted and applied at scoring time.

Because outcomes are rare, Box~2 prepared alternative training matrices on the \emph{development} slice only (evaluation splits were never modified): inverse-prevalence class weights, random oversampling (ROS), SMOTE-NC, Borderline-SMOTE, and ADASYN. For the main analysis, only the weights and SMOTE-NC variants were carried forward, and ROS and the Borderline-SMOTE/ADASYN fallbacks were retained for robustness in the appendix.

\subsection{Methodological Approach}\label{sec:method}

Each task was framed as a quarterly binary classification problem with non-overlapping time splits (Figure~\ref{fig:pipeline}, Boxes~3--5). The modeling approach follows best practices for predictive tasks with temporal dependence. Random cross-validation is known to induce leakage in time-indexed data because it disrupts the natural ordering and allows future information to influence training (Bergmeir et al., 2018). To address this, the panel is partitioned into non-overlapping development, holdout, and final windows, which aligns with rolling-origin and blocked forecasting methods shown to produce valid performance estimates \citep{Cerqueira2020ML, Roberts2017Ecography}. A Random Forest \citep{Breiman2001} is used as the baseline estimator in the model zoo and is compared with regularized logistic regression and gradient-boosting algorithms (XGBoost, LightGBM, CatBoost; see also \citealp{Sakhaee2025JEnvMan}), also used in complex operational settings \citep{Mohammadagha2025b}. This setup allows comparison between a linear baseline and flexible tree ensembles that capture non-linearities and interactions \citep{Sharifi2025DEL}.

\paragraph{Model development and selection (Box~3).}

For each dependent variable and each imbalance variant from Box~2, models were trained on the \emph{development} period using the persisted feature list for column alignment. Infinities were set to \texttt{NaN}, and any residual missingness was filled with development-fitted medians. Class weights or SMOTE-NC were applied only during development, and evaluation splits were never resampled or reweighted. Model selection prioritized PR-AUC as the primary criterion, with AUROC as a tiebreaker. Brier score and precision@K ($K\in{30,100,500}$) were also reported to quantify ranking usefulness. Evaluation was strictly out of time, with the \emph{final} window used for 12-month funding and the \emph{holdout} window used for 24-month patent growth and 36-month exits. The imbalance treatments follow established procedures for rare-event prediction. SMOTE and its variants are widely used to improve discrimination in settings with infrequent positive outcomes \citep{Chawla2002JAIR}, while inverse-prevalence weighting addresses small-sample bias in logistic-type models known to under-represent rare classes \citep{KingZeng2001PA}. These approaches are therefore suitable for sparse outcomes such as exits and near-term funding events.

\paragraph{Model analysis (Box~4).}

Interpretability and reliability were computed on the \emph{evaluable out-of-time split} (final or holdout, as applicable). Where feasible, \emph{TreeSHAP} was computed on the winning tree model using capped samples (and optional forest thinning) to obtain global importance (mean $|\text{SHAP}|$) and dependence plots. If TreeSHAP was infeasible due to memory or runtime, \emph{permutation importance} scored by average precision was used; for logistic baselines, coefficient magnitudes were reported. Each importance figure and table states the exact method used. Calibration was assessed with quantile-binned reliability curves and the \emph{Brier score}. Where curves showed optimism, it was noted that post hoc isotonic scaling can improve probability calibration without affecting ranking.

\paragraph{Out-of-sample prediction (Box~5).}

The winning specification for each outcome was applied to the most recent cohorts consistent with the horizon ($h{=}12/24/36$ months). Columns were aligned via the persisted feature list, development medians were applied unchanged, and no reweighting or resampling was used on evaluation data. Predicted probabilities $\hat{p}$, integer ranks, and percentiles were produced, with one row per organization after deduplication. Random seeds, file paths, and manifests were fixed to guarantee exact replication.

\section{Results} \label{sec:result}

\subsection{Descriptive Setup and First Signals}\label{sec:results-first}
This section reports diagnostics from the learning panel (2010-2023) and a univariate screen for each outcome. The panel is split by calendar time into a development sample (through 2019), a holdout sample (2020--2021), and a final test sample (2022--2023). By construction, only the 12-month funding outcome is evaluable in the final window; the 24-month patent and 36-month exit outcomes are evaluated on the holdout window. In the final window, the evaluable base rate for next financing within 12 months is 8.8\%.

Missingness is concentrated in two variables, \emph{days since last round} ($\approx$24\%) and \emph{firm age} ($\approx$6\%). Other predictors are essentially complete. Training rows typically contain at most one missing feature, which motivates the median imputation used in the pipeline (Refer to \S\ref{sec:data-processing}). Table~\ref{tab:correlations} summarizes pairwise correlations with the three outcomes.

The univariate screen is computed on development rows that are evaluable for each outcome to avoid look-ahead. Signed AUC is used (so values $>0.5$ indicate helpful discrimination regardless of direction), and PR-AUC magnitudes are reported; direction is indicated by the AUC sign (see Table~\ref{tab:predictors}).

\noindent\textbf{Next financing (12 months).} Signals related to recency and momentum dominate. \emph{Days since last round} loads negatively and shows meaningful stand-alone discrimination (signed AUC $\approx 0.69$; signed PR-AUC $\approx 0.27$). \emph{Age} is also negative (AUC $\approx 0.69$). Short-horizon momentum, \emph{rounds in the last four quarters} and \emph{funding in the last four quarters}, and accumulated exposure, \emph{cumulative investors} and \emph{cumulative rounds/raised}, load positively (AUCs $\approx 0.50$--$0.58$). Patent-stock variables have small negative associations.

\noindent\textbf{Patent growth (24 months).} The pattern is consistent with a change in stock rather than a short-run flow. Larger existing \emph{patent} and \emph{citation} stocks are negatively related to future increases, with relatively strong univariate lift (AUCs $\approx 0.68$--$0.70$). Younger firms are more likely to expand their portfolios (\emph{age}: AUC $\approx 0.70$, negative sign). Financing-activity variables carry small and mostly negative loadings.

\noindent\textbf{Exit (36 months).} Exit likelihood rises with maturity. \emph{Cumulative investors}, \emph{cumulative rounds}, and \emph{cumulative capital raised} are positive (AUCs $\approx 0.64$--$0.66$). Later-stage exposure, \emph{mid/late counts}, tilts positive (AUCs $\approx 0.54$--$0.55$). \emph{Days since last round} is negative (AUC $\approx 0.61$). Intellectual-property stocks show mild positive associations in the descriptive screen.

\noindent\textbf{Takeaway.} The screen aligns with economic intuition. Near-term fundraising is driven by recency and momentum, patent growth is mean-reverting with respect to existing IP stock and age, and exits move with accumulated financing maturity, with IP as a modest complement. These descriptive signals set priors for the model-based results that follow (\S\ref{sec:results-models}).


\subsection{Univariate Discrimination and Effect Sizes}\label{sec:results-univariate}
Signal strength for each predictor is quantified using \emph{signed} AUC and \emph{signed} PR-AUC computed on \emph{development} rows that are evaluable for each outcome to avoid look-ahead at panel edges.%

Table~\ref{tab:predictors} lists the five most informative features per outcome by signed PR-AUC; Table~\ref{tab:correlations} provides complementary correlations. Because several financing-history variables are correlated, for example cumulative rounds, investors, and capital, magnitudes are not additive and should be read as stand-alone lift only.

\paragraph{Next financing (12 months).}
\emph{Days since last round} and \emph{age} are the two dominant stand-alone signals (AUCs $\approx 0.69$; PR-AUCs $\approx 0.27$--$0.31$, both negative), followed by short-horizon momentum, \emph{rounds} and \emph{funding} in the last four quarters, and accumulated exposure, \emph{cumulative investors}. This confirms that recent activity and firm youth capture most of the univariate lift for near-term fundraising.

\paragraph{Patent growth (24 months).}
Mean reversion is first-order. Larger \emph{patent} and \emph{citation} stocks today imply lower odds of expansion within two years (AUCs $\approx 0.68$--$0.70$, negative). \emph{Age} enters with a strong negative sign, indicating that younger firms add patents, and \emph{cumulative rounds} contributes additional negative discrimination. Short-run funding momentum plays little role.

\paragraph{Exit (36 months).}
\emph{Cumulative investors}, \emph{cumulative rounds}, and \emph{cumulative capital raised} rank highest (AUCs $\approx 0.64$--$0.66$, positive), consistent with exits later in the financing lifecycle. \emph{Days since last round} remains negative (AUC $\approx 0.61$). Intellectual-property stocks show mild positive associations in the correlations (Table~\ref{tab:correlations}) but do not appear among the top-five predictors by signed PR-AUC (Table~\ref{tab:predictors}).

Overall, the diagnostics indicate that next-round fundraising is largely a recency and momentum phenomenon, patent growth is strongly mean-reverting and concentrated among younger firms, and exits are tied to accumulated financing maturity. In the next subsection, it is shown that multivariate models preserve these qualitative patterns while improving discrimination through non-linear interactions.


\subsection{Leakage-Safe Imputation and Feature Curation}\label{sec:results-impute}

The modeling feature space was completed by estimating a simple, leakage-safe imputation step on the \emph{development} period (2010--2019) and applying it unchanged to the \emph{holdout} (2020--2021) and \emph{final} (2022--2023) windows. The imputer is median-based, fit after replacing infinities with \texttt{NaN}, and it preserves all platform identifiers. To make missingness informative, explicit indicators were added for any feature with at least 10\% missing values in development, plus a forced flag for \emph{days since last round}. Consistent with the diagnostics in \S\ref{sec:results-first}, only \emph{days since last round} exceeds this threshold, while age is about 6\% and does not, so a binary \emph{days-since-last-round-is-missing} indicator is included in the modeling set.

Predictors were chosen using the univariate screen in \S\ref{sec:results-univariate}, supplemented by a small set of always-keep engineered variables that capture recency and capitalization. Table~\ref{tab:feature_subset} lists the features passed to the models for each outcome, saved in the persisted feature list and used for column alignment across splits, including the missingness indicator noted above. This procedure fixes the set used in \S\ref{sec:results-models} and ensures that preprocessing parameters are estimated once on development and then reused without modification, which avoids target leakage.


\subsection{Class Imbalance Handling}\label{sec:results-imbalance}

Class-imbalance variants were constructed only on the \emph{development} period (2010--2019), and the \emph{holdout} (2020--2021) or \emph{final} (2022--2023) samples were never modified. This guards against leakage by design. For each dependent variable, five training versions were created: a weights-only baseline using inverse-prevalence weights, random oversampling (ROS), SMOTE-NC, Borderline-SMOTE, and ADASYN. For SMOTE-NC, the categorical mask includes only the missingness indicator for \emph{days since last round}, and all other features are treated as numeric. All features in Table~\ref{tab:feature_subset} are numeric, and no industry or geography dummies are included. Consequently, the only categorical input to SMOTE-NC is the binary NA-indicator for \emph{days since last round}.

Table~\ref{tab:class_imbalance} reports class prevalence and whether each method runs on the panel. Training prevalence is moderate for funding and patent growth (17.6\% and 21.9\%) and low for exits (5.9\%). SMOTE-NC balances each outcome to roughly 50:50 as intended, as shown in the “SMOTE\_NC $N$ (balanced)” column. Borderline-SMOTE and ADASYN fail on this dataset due to residual \texttt{NaN} values in a small number of engineered features and therefore revert to ROS, which is not used in the main analysis.

To keep the main analysis focused and reproducible, two imbalance treatments per outcome were carried forward into model development: the \emph{weights-only} baseline and \emph{SMOTE-NC} balanced 50:50, exactly as summarized in the rightmost column of Table~\ref{tab:class_imbalance}. ROS and the Borderline-SMOTE or ADASYN fallbacks are used only for sensitivity checks outside the main text. This choice trades breadth for stability, predictable runtime, and a clean separation between preprocessing and modeling.

All outcomes are defined strictly forward from the quarter boundary with horizons $h\in{12,24,36}$ months. Rows too close to panel edges for full-horizon evaluation are marked non-evaluable and excluded from that task. Temporal splits are non-overlapping, with \emph{development} (2010-2019), \emph{holdout} (2020-2021), and \emph{final} (2022-2023). All preprocessing, infinite-to-\texttt{NaN}, NA indicators, and development-fitted medians, is estimated once on development and applied unchanged to later splits. Class weights and any resampling, SMOTE-NC or ROS, are confined to development, and evaluation splits are never reweighted or resampled.


\subsection{Model Development and Selection}\label{sec:results-models}

For each dependent variable, five model families were trained: logistic regression, Random Forest, XGBoost, LightGBM, and CatBoost. Each was estimated under two leakage-safe imbalance variants from \S\ref{sec:results-imbalance}, inverse-prevalence \emph{weights} and \emph{SMOTE-NC}.

Training uses the \emph{development} period (2010--2019), with features and medians fixed in \S\ref{sec:results-impute} through the persisted feature list and development-fitted medians. Evaluation is strictly out of time, the \emph{final} window (2022--2023) when the label is observable, otherwise the \emph{holdout} window (2020--2021). Headline metrics are PR-AUC as the primary criterion, AUROC, and Brier score, with sample sizes and base rates shown alongside results. By construction, only the 12-month funding outcome is evaluable in the final window. The 24-month patent and 36-month exit outcomes are evaluated on the holdout window.

Table \ref{tab:leaderboard} is the single source of truth for results.
(i) \emph{Funding (12 months).} On the final window, the LightGBM model trained with weights attains AUROC $=0.817$, PR-AUC $=0.220$, and Brier $=0.144$.
(ii) \emph{Patent growth (24 months).} With only the holdout window evaluable, the weights-only Random Forest delivers PR-AUC $=0.631$ and AUROC $=0.921$ (Brier $=0.074$).
(iii) \emph{Exit (36 months).} On holdout, the weights-only Random Forest yields PR-AUC $=0.559$ and AUROC $=0.872$ (Brier $=0.032$).

Within the pre-specified variants, weights and SMOTE-NC, Random Forest dominates gradient-boosting and logistic baselines on patent growth and exits, while \emph{LightGBM with inverse-prevalence weights} is selected for near-term fundraising. These winners are carried forward to \S\ref{sec:results-shap} for interpretability and calibration diagnostics and to \S\ref{sec:results-targets} for out-of-time scoring summaries.

\subsection{Model Analysis and Interpretability}\label{sec:results-shap}

\begin{figure}[t]
  \centering
  \includegraphics[width=.82\linewidth]{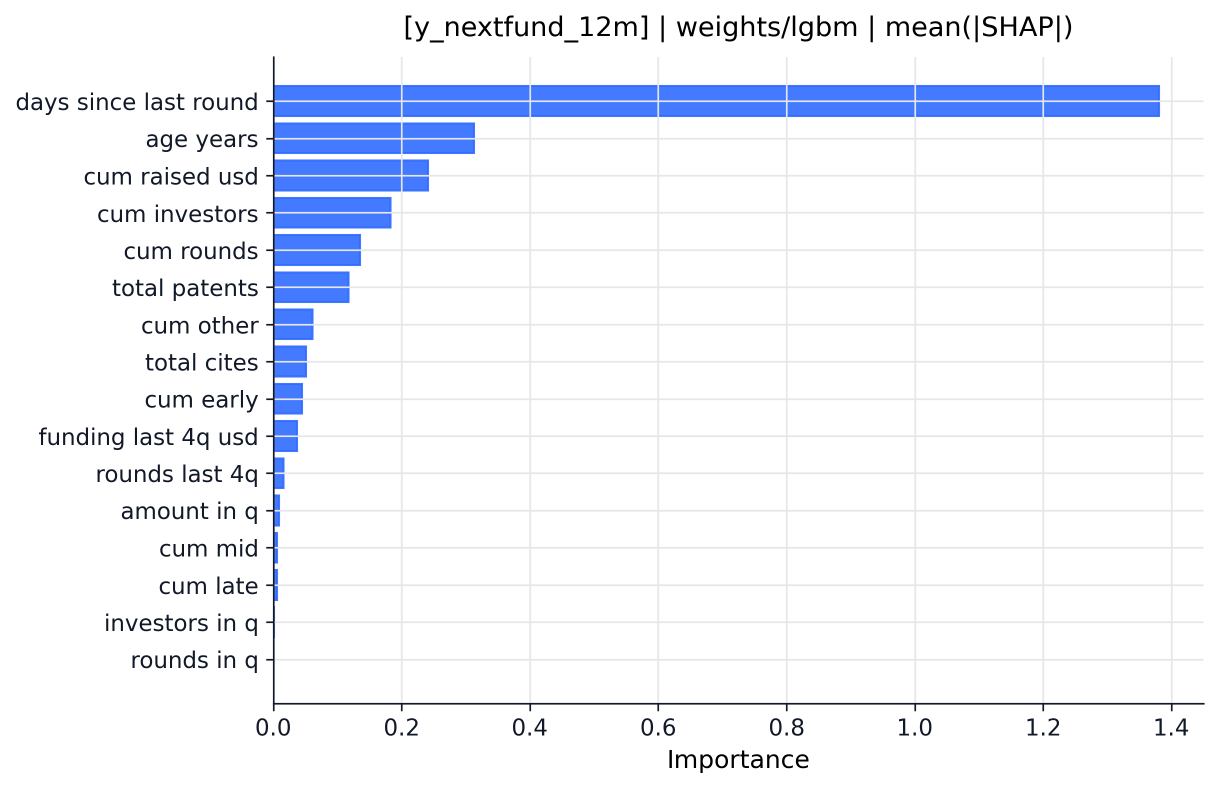}
  \caption{Funding (12m), LightGBM trained with inverse-prevalence weights. Bars show global feature importance measured by mean absolute SHAP values ($\text{mean}(|SHAP|)$) on the final evaluation window. Financing recency and firm age dominate, followed by cumulative capital raised and investor breadth.}
  \label{fig:imp_fund}
\end{figure}

\begin{figure}[t]
  \centering
  \includegraphics[width=.72\linewidth]{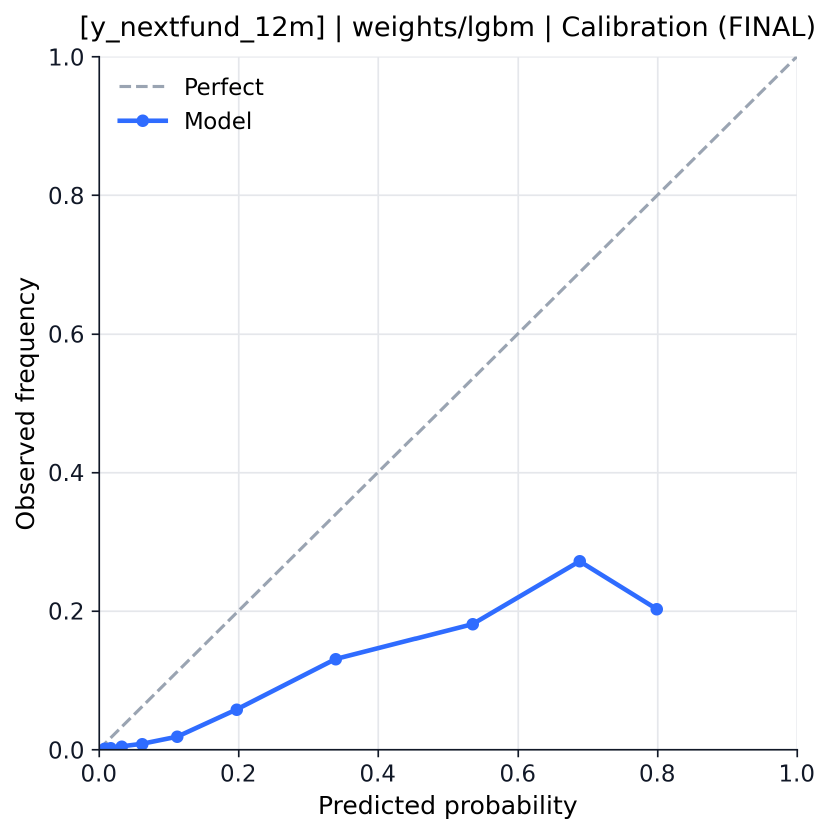}
  \caption{Funding (12m), LightGBM trained with inverse-prevalence weights. Quantile-binned calibration curve on the \emph{final} window. The model shows optimism at higher predicted probabilities relative to the $45^\circ$ reference line, suggesting that isotonic recalibration could improve probability calibration without affecting ranking.}
  \label{fig:cal_fund}
\end{figure}

\begin{figure}[t]
  \centering

  \begin{subfigure}[t]{0.48\linewidth}
    \centering
    \includegraphics[width=\linewidth]{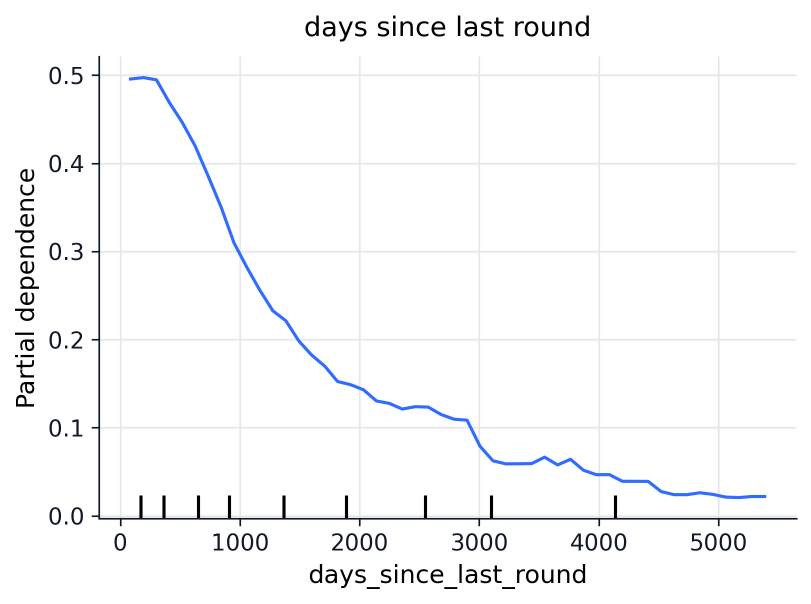}
    \caption{Days since last round}
  \end{subfigure}\hfill
  \begin{subfigure}[t]{0.48\linewidth}
    \centering
    \includegraphics[width=\linewidth]{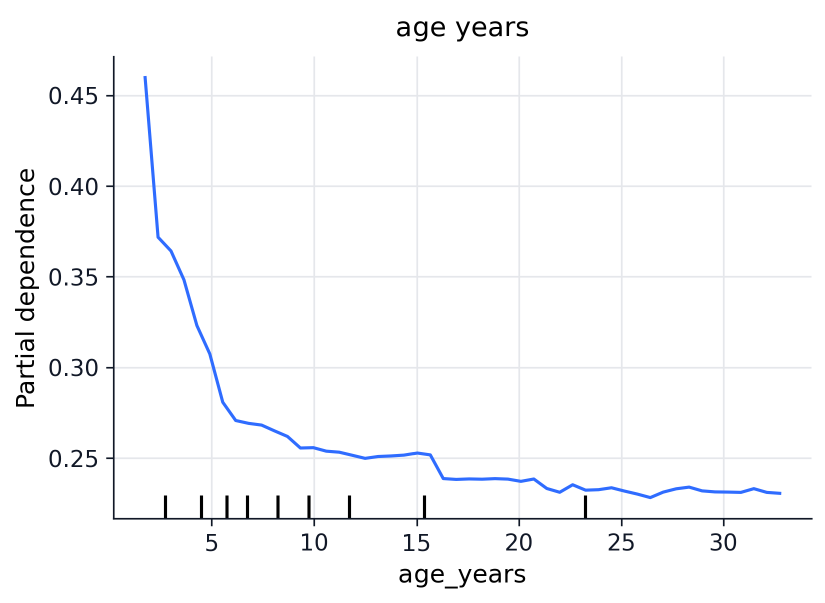}
    \caption{Age (years)}
  \end{subfigure}

  \vspace{0.7em}

  \begin{subfigure}[t]{0.48\linewidth}
    \centering
    \includegraphics[width=\linewidth]{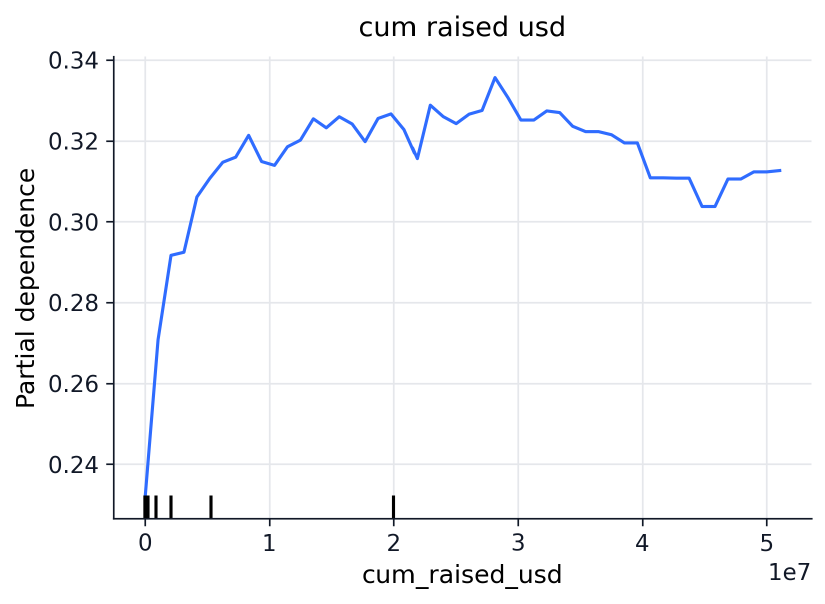}
    \caption{Cumulative \$ raised}
  \end{subfigure}\hfill
  \begin{subfigure}[t]{0.48\linewidth}
    \centering
    \includegraphics[width=\linewidth]{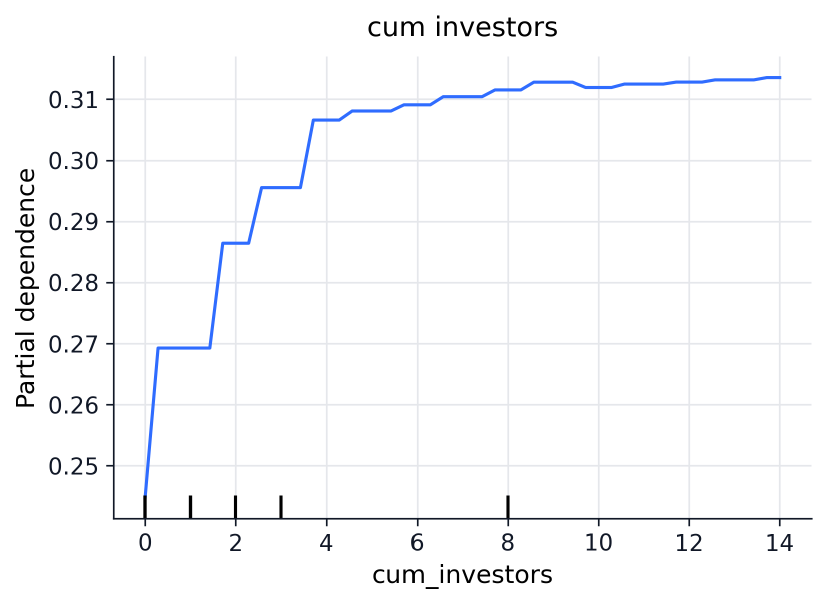}
    \caption{Cumulative investors}
  \end{subfigure}

  \vspace{0.7em}

  \begin{subfigure}[t]{0.48\linewidth}
    \centering
    \includegraphics[width=\linewidth]{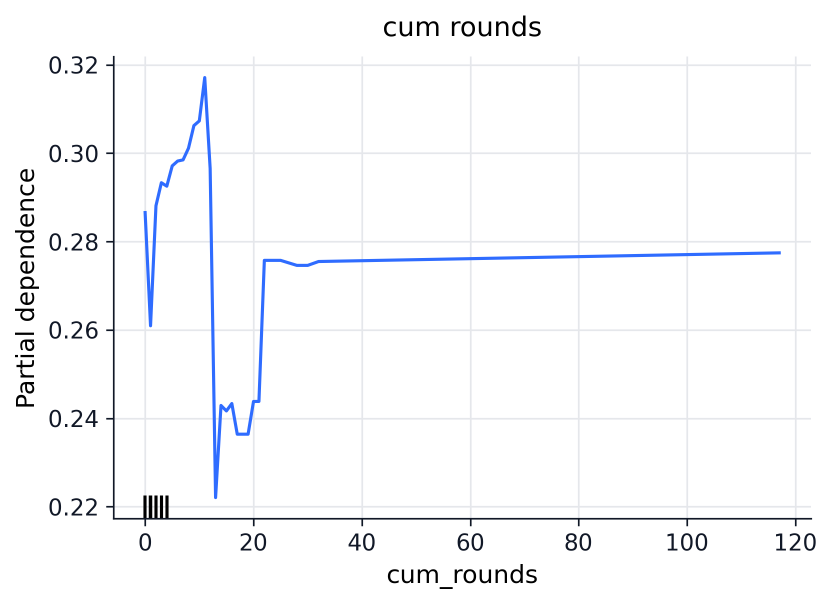}
    \caption{Cumulative rounds}
  \end{subfigure}\hfill
  \begin{subfigure}[t]{0.48\linewidth}
    \centering
    \includegraphics[width=\linewidth]{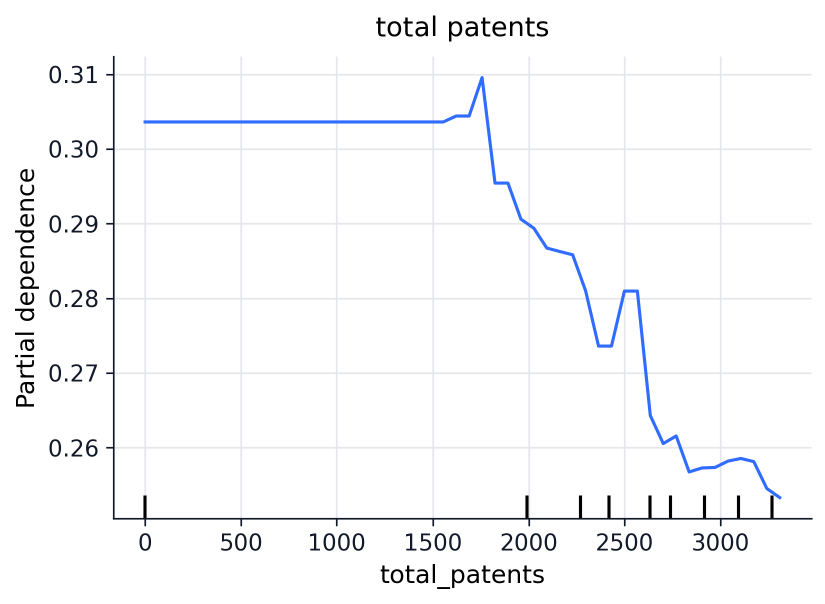}
    \caption{Total patents}
  \end{subfigure}

  \caption{Funding (12m), LightGBM trained with inverse-prevalence weights. Partial dependence plots on the \emph{final} window for the top six predictors. The probability of next-round funding decreases with time since the last round and firm age, and rises with cumulative funding, investor breadth, and round count. Higher patent stock shows a mild negative effect. \emph{Note:} Uncalibrated probabilities appear overconfident at high scores, but ranking remains unaffected.}
  \label{fig:pdp_fund}
\end{figure}

\begin{figure}[t]
  \centering
  \includegraphics[width=.82\linewidth]{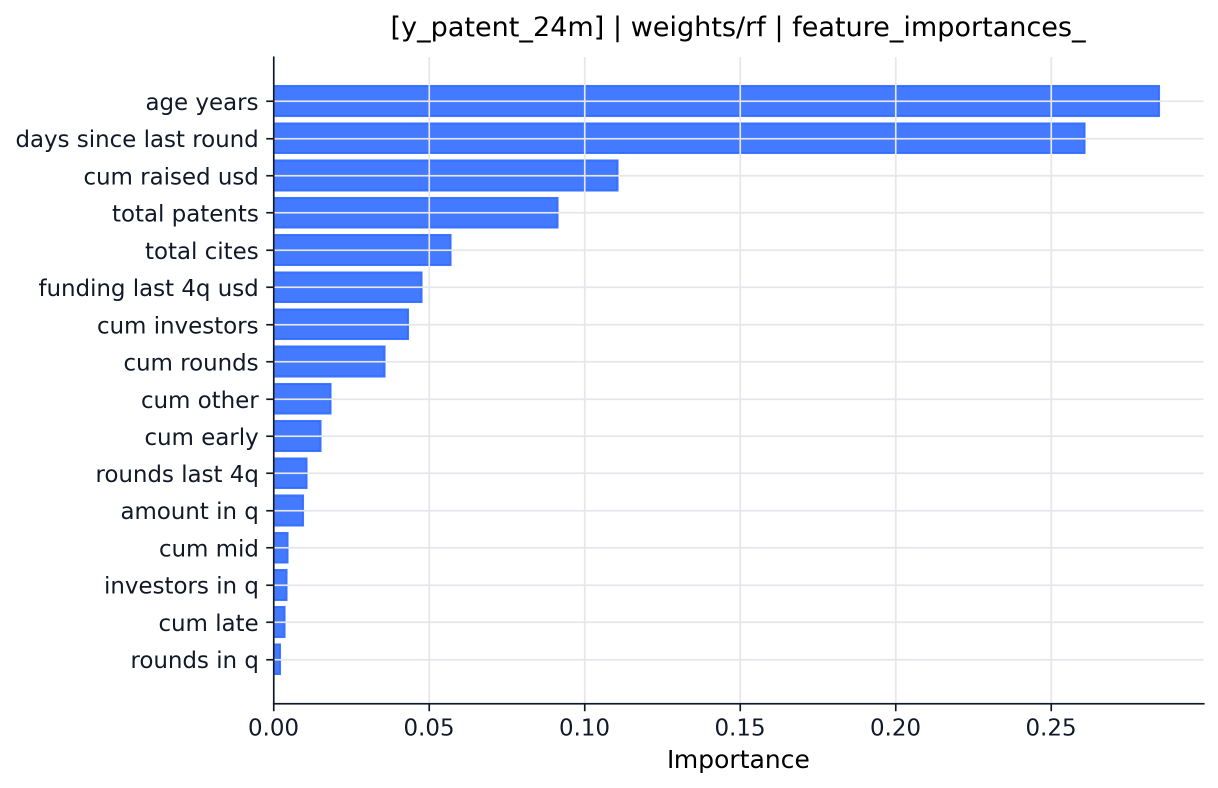}
  \caption{Patent growth (24m), Random Forest trained with inverse-prevalence weights. Bars show global feature importance based on mean decrease in impurity (MDI, Gini). Firm age, time since last round, and cumulative capital raised dominate the ranking, followed by patent and citation stocks. \emph{Note:} MDI can overemphasize high-cardinality or correlated predictors.}
  \label{fig:imp_pat}
  \footnotesize
\end{figure}

\begin{figure}[t]
  \centering
  \includegraphics[width=.72\linewidth]{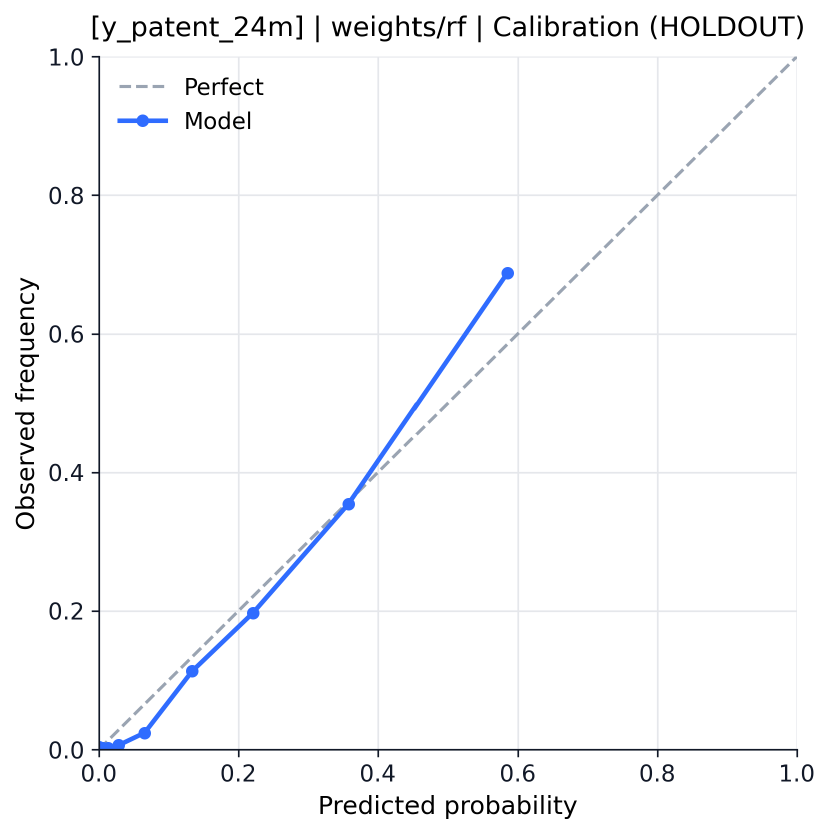}
  \caption{Patent growth (24m), Random Forest trained with inverse-prevalence weights. Quantile-binned calibration curve on the \emph{holdout} window. The model aligns closely with the $45^\circ$ reference line, showing well-calibrated predicted probabilities across most of the range.}
  \label{fig:cal_pat}
\end{figure}

\begin{figure}[t]
  \centering

  \begin{subfigure}[t]{0.48\linewidth}
    \centering
    \includegraphics[width=\linewidth]{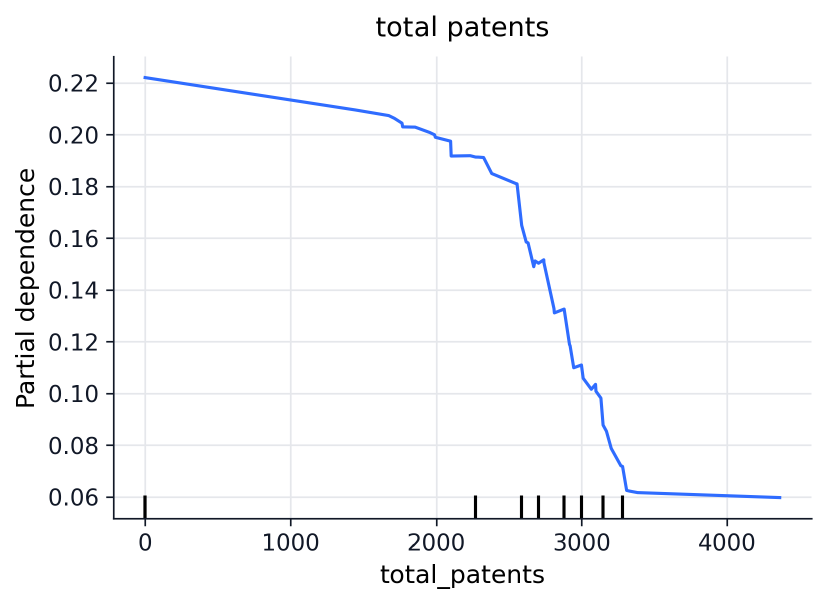}
    \caption{Total patents}
  \end{subfigure}\hfill
  \begin{subfigure}[t]{0.48\linewidth}
    \centering
    \includegraphics[width=\linewidth]{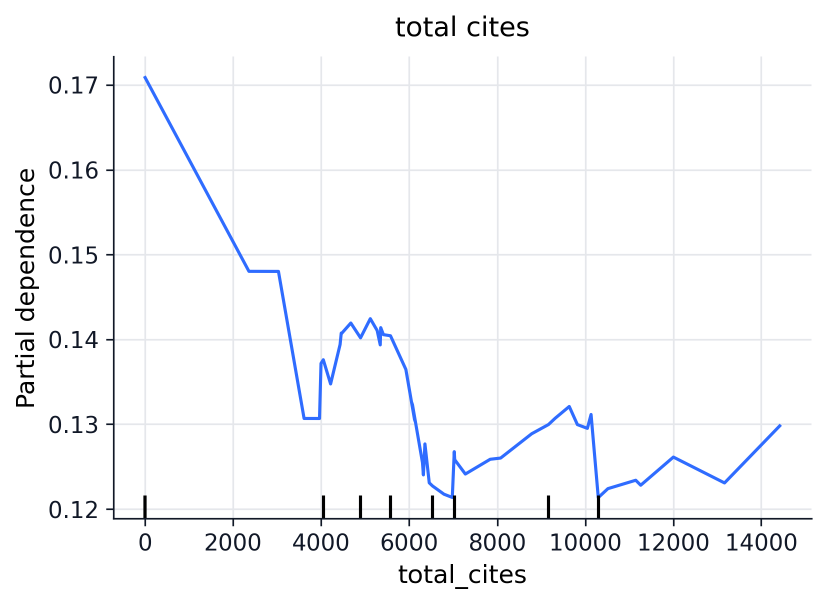}
    \caption{Total citations}
  \end{subfigure}

  \vspace{0.7em}

  \begin{subfigure}[t]{0.48\linewidth}
    \centering
    \includegraphics[width=\linewidth]{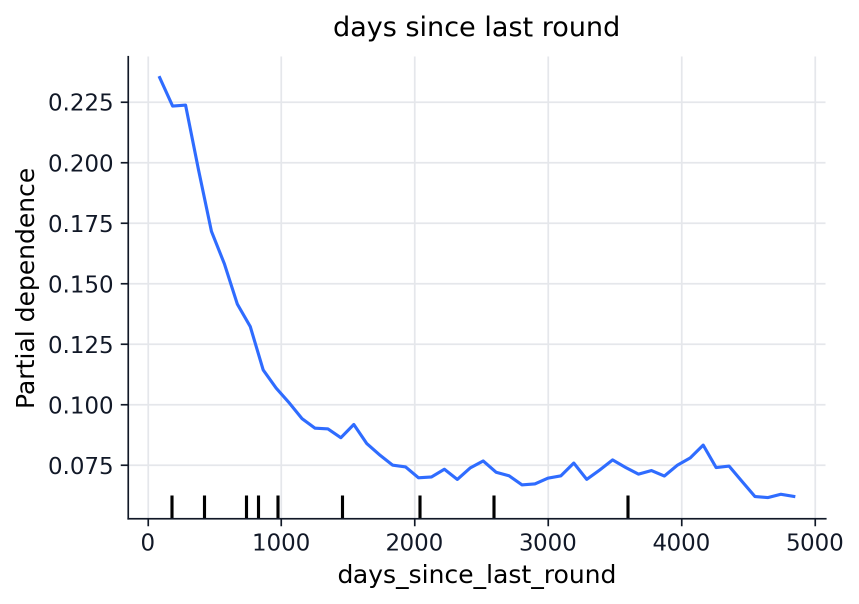}
    \caption{Days since last round}
  \end{subfigure}\hfill
  \begin{subfigure}[t]{0.48\linewidth}
    \centering
    \includegraphics[width=\linewidth]{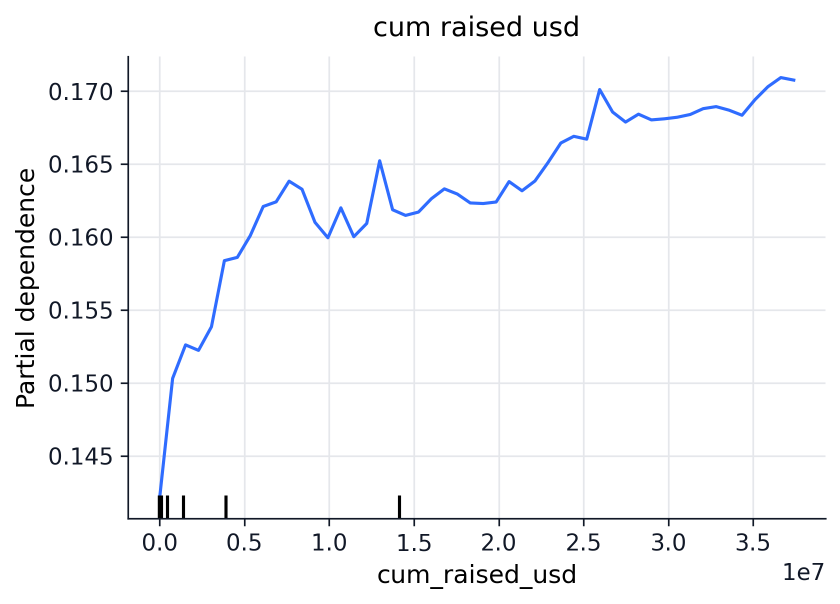}
    \caption{Cumulative \$ raised}
  \end{subfigure}

  \vspace{0.7em}

  \begin{subfigure}[t]{0.48\linewidth}
    \centering
    \includegraphics[width=\linewidth]{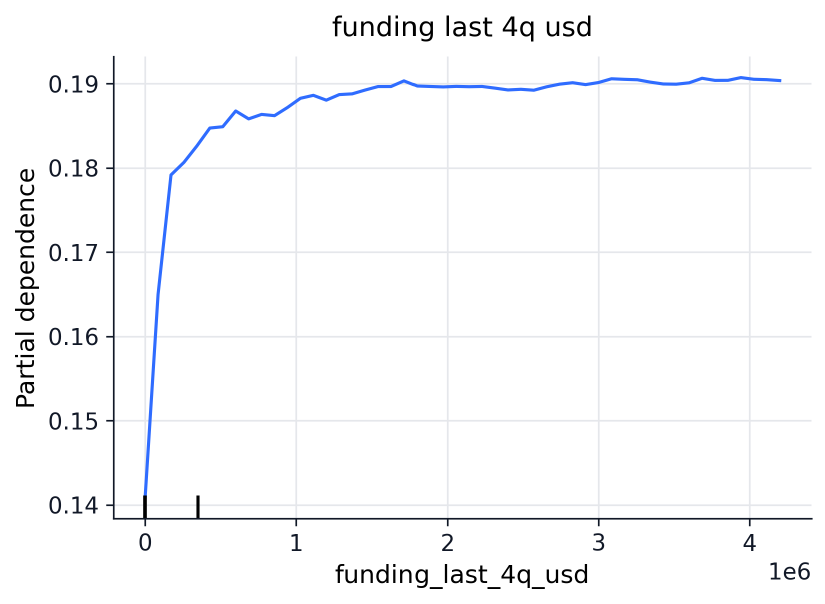}
    \caption{Funding last 4q}
  \end{subfigure}\hfill
  \begin{subfigure}[t]{0.48\linewidth}
    \centering
    \includegraphics[width=\linewidth]{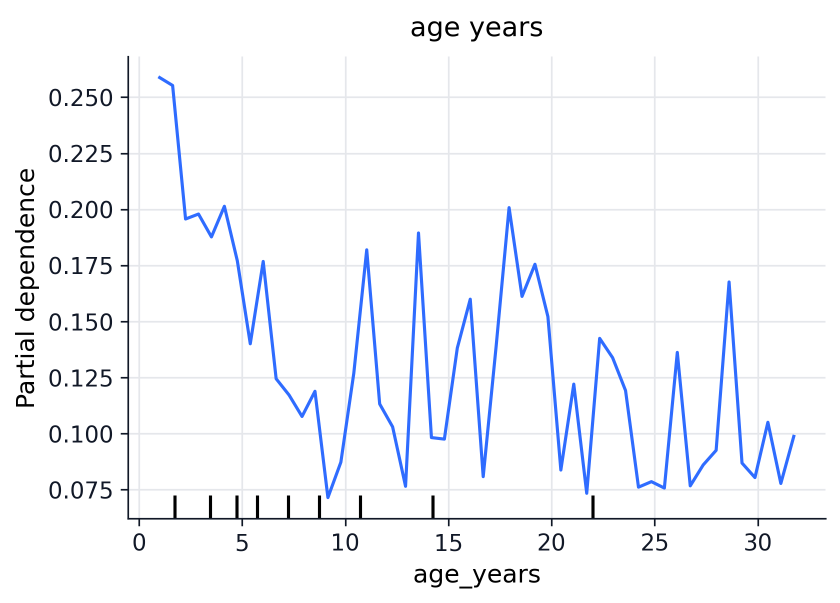}
    \caption{Age (years)}
  \end{subfigure}

  \caption{Patent growth (24m), Random Forest trained with inverse-prevalence weights. Partial dependence plots on the \emph{holdout} window for six key predictors. The probability of patent expansion decreases with existing patent and citation stocks and with time since the last round, and increases with cumulative and recent funding. Younger firms show higher predicted probabilities, consistent with maturity and mean-reversion effects.}
  \label{fig:pdp_pat}
\end{figure}

\begin{figure}[t]
  \centering
  \includegraphics[width=.82\linewidth]{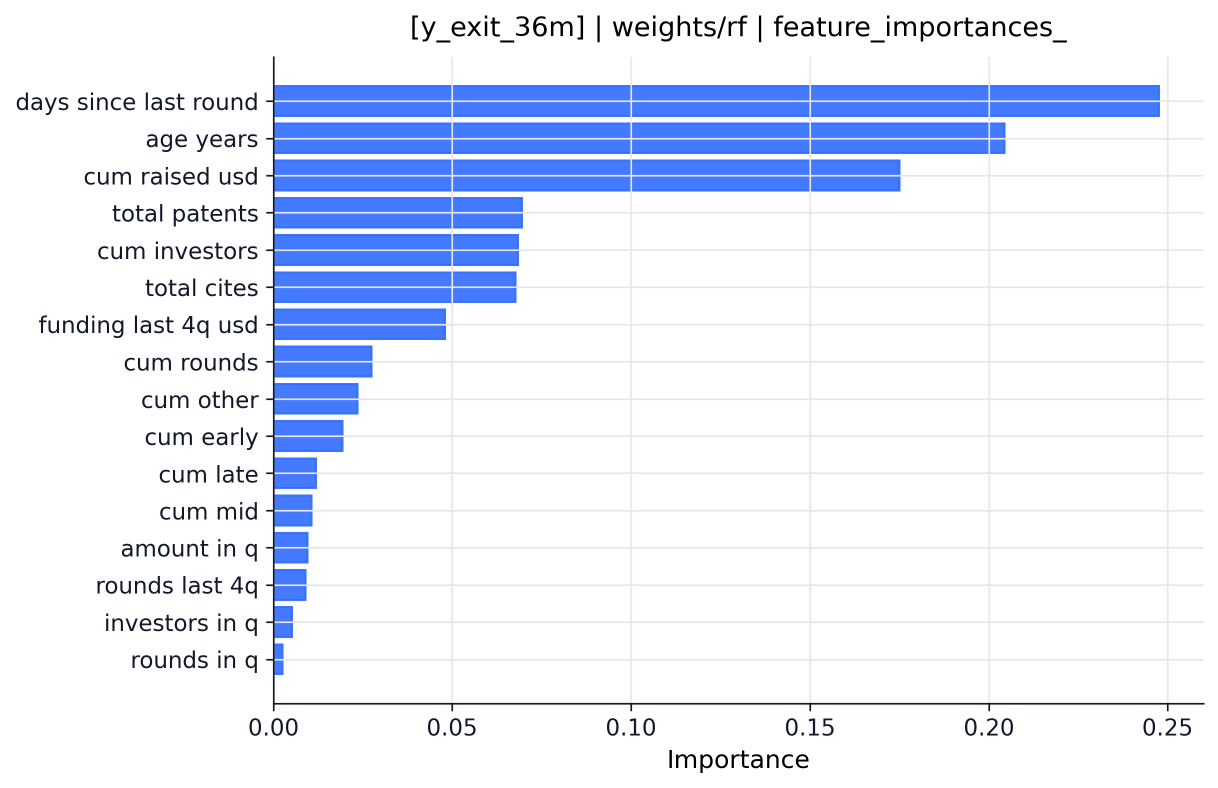}
  \caption{Exit (36m), Random Forest trained with inverse-prevalence weights. Bars show global feature importance based on mean decrease in impurity (MDI, Gini). Exit predictions are driven by financing maturity measures such as time since last round, firm age, and cumulative capital raised, with intellectual property variables contributing positively but secondarily. \emph{Note:} MDI may overweight features with higher cardinality or correlation.}
  \label{fig:imp_exit}
  \footnotesize
\end{figure}

\begin{figure}[t]
  \centering
  \includegraphics[width=.72\linewidth]{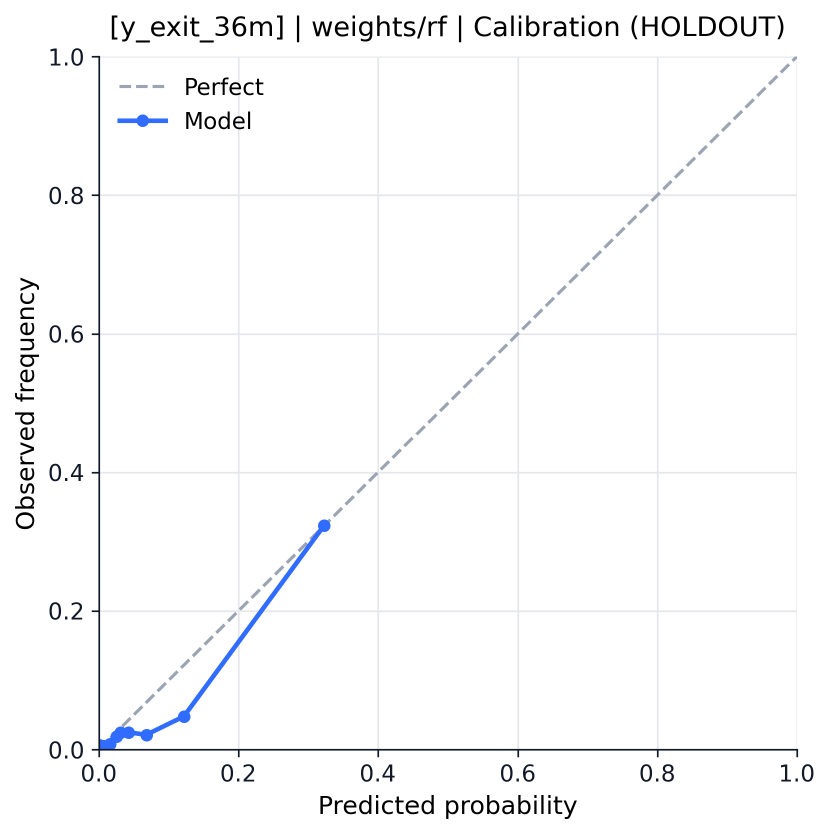}
  \caption{Exit (36m), Random Forest trained with inverse-prevalence weights. Quantile-binned calibration curve on the \emph{holdout} window. The model follows the $45^\circ$ reference line closely at low and mid probability ranges, indicating good calibration, with mild optimism at higher predicted probabilities.}
  \label{fig:cal_exit}
\end{figure}

\begin{figure}[t]
  \centering

  \begin{subfigure}[t]{0.48\linewidth}
    \centering
    \includegraphics[width=\linewidth]{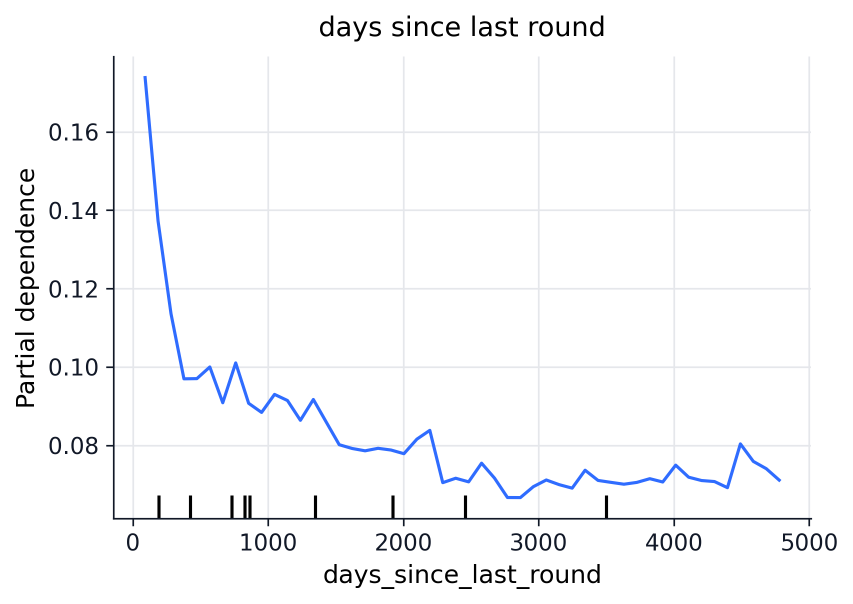}
    \caption{Days since last round}
  \end{subfigure}\hfill
  \begin{subfigure}[t]{0.48\linewidth}
    \centering
    \includegraphics[width=\linewidth]{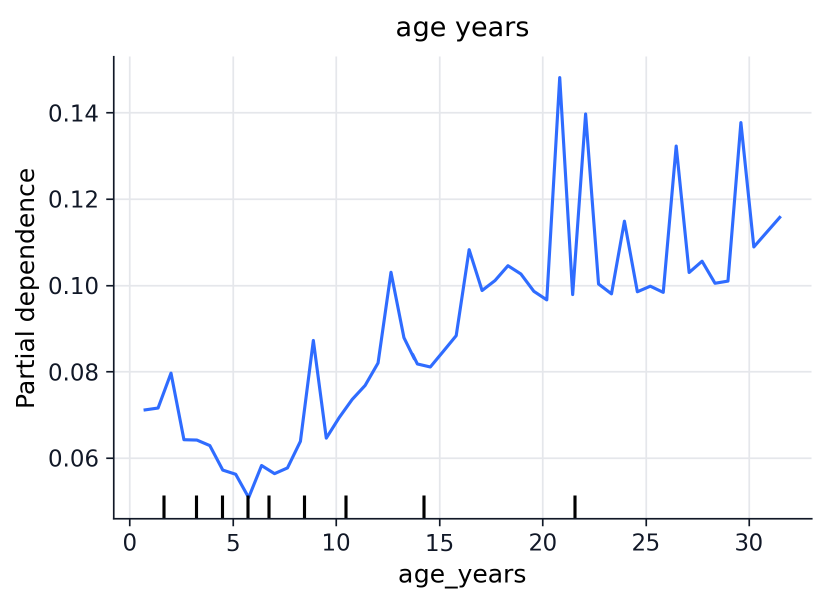}
    \caption{Age (years)}
  \end{subfigure}

  \vspace{0.7em}

  \begin{subfigure}[t]{0.48\linewidth}
    \centering
    \includegraphics[width=\linewidth]{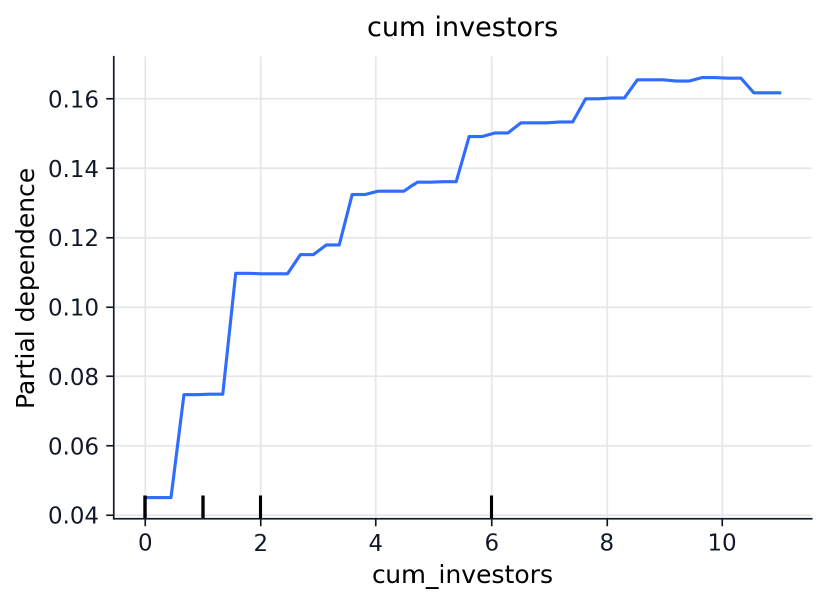}
    \caption{Cumulative investors}
  \end{subfigure}\hfill
  \begin{subfigure}[t]{0.48\linewidth}
    \centering
    \includegraphics[width=\linewidth]{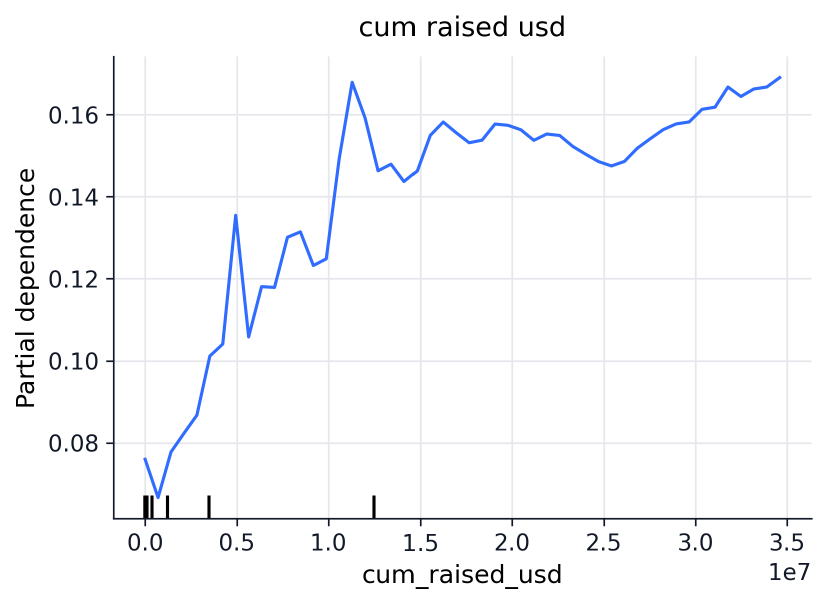}
    \caption{Cumulative \$ raised}
  \end{subfigure}

  \vspace{0.7em}

  \begin{subfigure}[t]{0.48\linewidth}
    \centering
    \includegraphics[width=\linewidth]{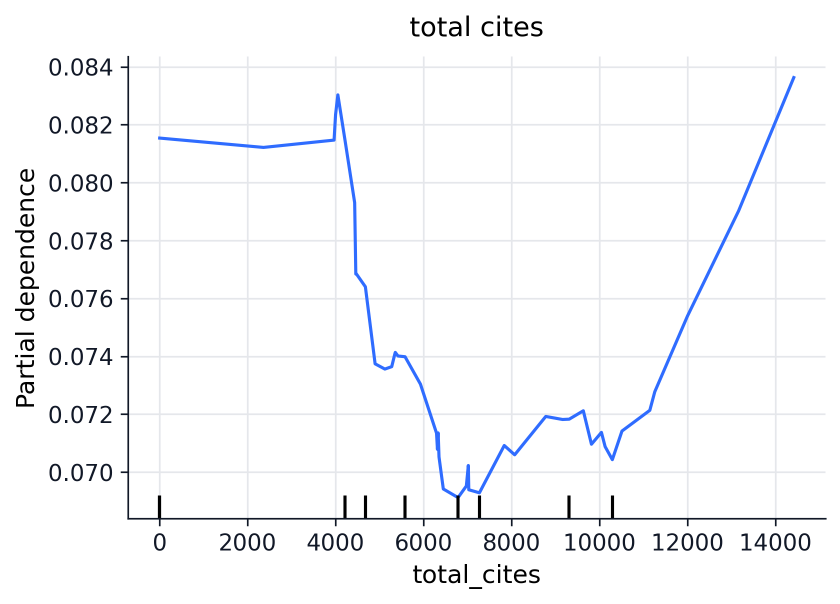}
    \caption{Total citations}
  \end{subfigure}\hfill
  \begin{subfigure}[t]{0.48\linewidth}
    \centering
    \includegraphics[width=\linewidth]{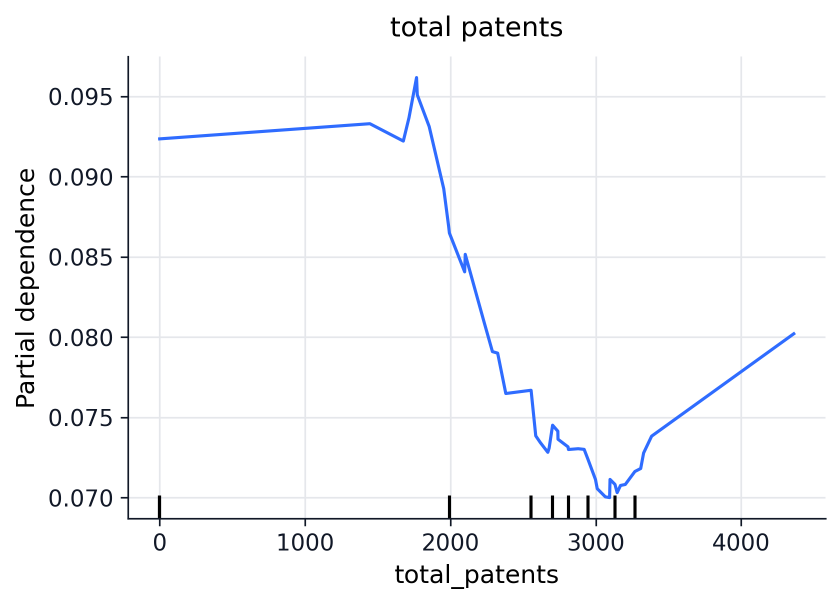}
    \caption{Total patents}
  \end{subfigure}

  \caption{Exit (36m), Random Forest trained with inverse-prevalence weights. Partial dependence plots on the \emph{holdout} window for six key predictors. Exit likelihood increases with cumulative investors and total capital raised, and decreases with time since last round. Age and intellectual property variables show weaker but consistent positive associations.}
  \label{fig:pdp_exit}
\end{figure}

The winning specification for each outcome is evaluated using three diagnostics: global feature importance; reliability via calibration curves and the Brier score; and partial dependence for the most influential predictors. All diagnostics are strictly out of time. Summary performance metrics are reported in Table~\ref{tab:leaderboard}, and interpretability artifacts and reliability are visualized in Figures~\ref{fig:imp_fund}--\ref{fig:pdp_exit}.

For the funding winner, LightGBM with weights, \emph{TreeSHAP} is computed on capped samples. For the patent and exit winners, Random Forest with weights, \emph{feature\_importances\_} based on Gini importance and, where applicable, permutation importance scored by average precision are reported, and partial dependence is provided for top features. Calibration is assessed with quantile-binned reliability curves and the Brier score. Where curves show optimism, post hoc isotonic recalibration is recommended, with ranking unaffected.

\paragraph{Funding within 12 months (winner: LGBM, weights).}
On the \emph{final} window, AUROC $=0.817$, PR\mbox{-}AUC $=0.220$, and Brier $=0.144$ (Table~\ref{tab:leaderboard}). The SHAP profile in Figure~\ref{fig:imp_fund} is led by \emph{days since last round} and \emph{age}, followed by \emph{cumulative capital raised} and \emph{cumulative investors}. Partial dependence in Figure~\ref{fig:pdp_fund} indicates that the probability of another round \emph{decreases} as time since the last round increases and with firm age, and \emph{increases} with recent financing intensity and with cumulative funding and investor breadth, with concave gains at higher levels. The calibration curve in Figure~\ref{fig:cal_fund} shows optimism across most of the score range, strongest at higher predicted probabilities, so post hoc isotonic scaling would improve probability calibration without affecting ranking.

\paragraph{Patent growth within 24 months (winner: RF, weights).}
On the \emph{holdout} window, AUROC $=0.921$, PR\mbox{-}AUC $=0.631$, and Brier $=0.074$ (Table~\ref{tab:leaderboard}). Global importance, Figure~\ref{fig:imp_pat}, identifies \emph{firm age}, \emph{time since last round}, and \emph{cumulative capital raised} as the dominant predictors, with patent and citation stocks contributing at secondary magnitudes. The partial-dependence panels in Figure~\ref{fig:pdp_pat} show economically plausible associations: predicted patent growth \emph{declines} with existing IP stock and with time since last funding, \emph{increases} with cumulative and recent funding intensity, and is \emph{higher for younger firms}. The reliability curve in Figure~\ref{fig:cal_pat} is close to the $45^\circ$ line across most of the range, with slight underconfidence in the top bin. Isotonic recalibration is advisable if calibrated probabilities are required.

\paragraph{Exit within 36 months (winner: RF, weights).}
On the \emph{holdout} window, AUROC $=0.872$, PR\mbox{-}AUC $=0.559$, and Brier $=0.032$ (Table~\ref{tab:leaderboard}). Global importance, Figure~\ref{fig:imp_exit}, indicates that \emph{time since last round}, \emph{age}, and \emph{cumulative capital raised} are the leading predictors, with investor breadth, cumulative rounds, and IP stocks contributing at secondary levels. Partial dependence in Figure~\ref{fig:pdp_exit} shows that exit likelihood \emph{decreases} as time since the last round increases and \emph{increases} with cumulative capital raised and investor breadth; it also \emph{increases with age}. IP variables exhibit weaker, partly non-monotonic but generally positive associations at higher levels. Calibration in Figure~\ref{fig:cal_exit} is close to well calibrated at low to mid scores with mild optimism in the right tail, so scores are best used for ranking unless probabilities are re-calibrated.

\medskip

\noindent\textbf{Notes on importance.} For the funding tree model, LightGBM, \emph{TreeSHAP} is used. For the patent and exit winners, Random Forest, \emph{feature\_importances\_} based on Gini importance are reported, and where permutation or average-precision scoring is used, the figure and caption state it explicitly. Across outcomes, financing recency and momentum are consistently informative. Innovation stock is strongly predictive of subsequent patent growth and only modestly informative for near-term fundraising. Exit prediction moves with accumulated financing maturity. Ranking is useful, and calibrated probabilities can be obtained via post hoc isotonic scaling when needed.


\subsection{Out-of-time scoring and ranked target lists}\label{sec:results-targets}

This step applies the winning specification for each outcome to the most recent cohort for which the prediction horizon is fully evaluable. Columns are aligned using the persisted feature list, infinities are replaced with \texttt{NaN}, and missing entries are filled with development-period medians. No resampling or reweighting is applied to evaluation data. Predictions $\hat p$ are converted to integer ranks and percentiles. A single row per organization is retained after deduplication by keeping the highest score and the most recent quarter when ties occur.
Table~\ref{tab:scoring_outputs} summarizes winners, selection splits, and scored cohorts. Performance metrics used to select the winner come from Table \ref{tab:leaderboard}, with capacity-style checkpoints in Table~\ref{tab:precision_at_k}, and are not re-estimated here.

The ranked lists are intended for screening and benchmarking. Given that Section~\ref{sec:results-shap} documents discrimination and calibration, only a manifest of the scored cohorts is reported in the main text. If calibrated probabilities are required for decision-making, isotonic recalibration can be applied post hoc without changing ranking.

The scoring step translates the validated models into actionable target lists. Cohort choices reflect the horizon length, 2022 for financing within 12 months, 2021 for patent growth within 24 months, and 2020 for exits within 36 months. These outputs enable practical prioritization while preserving the leakage controls defined in Section~\ref{sec:method}.

\section{Conclusion} \label{sec:conclusion}

This paper examined whether near-term funding, medium-horizon patent growth, and longer-horizon exits for startups can be predicted in a leakage-safe and interpretable way using routinely available information. A firm-quarter panel was built from Crunchbase and merged to USPTO assignee data. A pipeline was designed to train only on a development window and to evaluate strictly out of time on non-overlapping holdout and final windows. Model selection emphasized ranking performance under class imbalance, and post-estimation diagnostics paired discrimination with interpretability and reliability.

Three findings emerge. First, patent expansion over a two-year horizon was highly predictable at the firm-quarter level on the holdout window. A Random Forest with inverse-prevalence weights delivered strong discrimination, and interpretation based on feature importance and partial dependence indicated maturity and financing recency as central drivers, consistent with views of patents as protective assets and as credible signals that ease financing frictions \citep{Conti2013JOIE,HsuZiedonis2013SMJ,FarreMensa2020JF}. Second, next funding within 12 months was best captured by a weighted LightGBM on the final window. Explanations indicate pronounced roles for time since last round, age, and cumulative financing intensity, consistent with evidence that venture outcomes reflect momentum in investor attention and organizational maturity \citep{Gompers2010JFE,KortumLerner2000RAND}. Third, exit predictions at 36 months again favored Random Forests with weights, and the most important predictors relate to the depth and breadth of prior financing, with intellectual property stocks contributing positively but secondarily, consistent with trajectories in which capital structure and investor networks shape eventual liquidity events. An additional implication concerns the role of patent-based innovation signals. Studies in innovation economics show that early patent strength, novelty, or citation potential can forecast future technological growth \citep{Kelly2021AERI} and can influence investor expectations in high-uncertainty settings. The importance of firm age, mean reversion, and recent financing in our patent-growth models aligns with these findings and suggests that patent-stock expansions are driven by both maturity and evolving research intensity. The findings also align with recent work showing that combining interpretable ML with structured innovation and financing signals can meaningfully improve forecasting in entrepreneurial settings \citep{Kim2023TFSC,Maarouf2024EJOR}.

Methodologically, the paper contributes an interpretable ML template for innovation finance. Development-only preprocessing was used, including infinite-to-\texttt{NaN}, missing flags, and median imputation. Persisted feature lists ensured column alignment, and clean temporal splits avoided look-ahead. The comparative model zoo across linear and tree-based learners under inverse-prevalence weights and SMOTE-NC follows best practice in explainable prediction \citep{Breiman2001,Kim2024,Mohammadagha2025c}. SHAP and partial dependence provided narratives that link scores to economically meaningful margins \citep{LundbergLee2017NeurIPS}. Calibration analysis highlights when isotonic scaling is advisable before probabilities are used for decision thresholds, while ranking use cases remain robust.

Limitations suggest clear extensions. Platform coverage and survivorship in Crunchbase can introduce selection and measurement error \citep{Dalle2017OECD,Mohammadagha2025a}. Expanding sources or auditing against administrative records would test robustness. The focus here is on structured variables. Incorporating text from patents, company descriptions, and filings could enrich signals, provided interpretability is preserved. Although strict time splits mitigate leakage, regime shifts, including post-pandemic funding cycles, can alter base rates and feature distributions. Formal drift monitoring and periodic recalibration would support deployment. Finally, outcomes are coarse, defined as any funding, any patent growth, or any exit. Future work could model timing with survival methods, explore heterogeneity by sector or technology class \citep{Kogan2017QJE}, and evaluate counterfactual policy experiments under explainable learners.

In sum, leakage-safe and interpretable ML was shown to yield ranking-useful predictions for startup funding, patenting, and exits, while recovering mechanisms that align with established theory and demonstrates the usefulness of interpretable ML frameworks across high-dimensional real-world datasets (e.g., \citet{Sholehrasa2025}). The pipeline’s emphasis on transparency, through time-clean evaluation and economically legible explanations, makes it suitable for scholarly replication, investor screening, and policy monitoring. Extending the framework with richer data modalities and drift-aware maintenance promises greater practical relevance for innovation-finance research.

\clearpage
\singlespacing
\setlength\bibsep{0pt}

\addcontentsline{toc}{section}{References}



\clearpage
\onehalfspacing

\section*{Figures} \label{sec:fig}
\addcontentsline{toc}{section}{Figures}

\vspace{0.3in}

\FloatBarrier


\FloatBarrier

\FloatBarrier

\FloatBarrier


\FloatBarrier

\FloatBarrier

\FloatBarrier


\FloatBarrier

\FloatBarrier

\FloatBarrier


\clearpage
\section*{Tables} \label{sec:tab}
\addcontentsline{toc}{section}{Tables}
\vspace{0.3in}
\FloatBarrier

\begin{table}[htbp]
\centering
\caption{Pairwise correlations between selected firm-level features and three startup outcomes on the development panel (2010–2019). Correlations are computed at the firm–quarter level after preprocessing. Positive values indicate features associated with a higher likelihood of the outcome, while negative values indicate inverse relationships. Patterns reflect financing recency and momentum for funding, mean reversion in patenting, and maturity effects for exits. \emph{Correlations are linear and may differ in sign from partial dependence shapes when effects are non-linear or interact with other variables (e.g., age in Exit)}.
}
\label{tab:correlations}
\resizebox{\textwidth}{!}{%
\begin{tabular}{l
                S[table-format=-1.3]
                S[table-format=-1.3]
                S[table-format=-1.3]}
\toprule
{Feature} & {Next funding (12m)} & {Patent growth (24m)} & {Exit (36m)} \\
\midrule
Total patents                     & -0.049 & -0.294 &  0.078 \\
Total citations                   & -0.056 & -0.252 &  0.072 \\
Days since last round             & -0.222 & -0.195 & -0.071 \\
Cumulative investors              &  0.114 & -0.039 &  0.162 \\
Rounds in last four quarters      &  0.163 &  0.006 &  0.078 \\
Cumulative rounds                 &  0.065 & -0.138 &  0.132 \\
Cumulative mid-stage rounds       &  0.064 & -0.007 &  0.126 \\
Cumulative late-stage rounds      &  0.052 & -0.004 &  0.130 \\
Cumulative early-stage rounds     &  0.051 & -0.097 &  0.067 \\
Cumulative other-stage rounds     &  0.025 & -0.123 &  0.074 \\
Capital raised in last 4 quarters &  0.028 & -0.002 &  0.040 \\
Cumulative capital raised (USD)   &  0.029 & -0.013 &  0.068 \\
Capital raised in current quarter &  0.014 & -0.002 &  0.022 \\
Rounds in current quarter         &  0.080 &  0.005 &  0.036 \\
Investors in current quarter      &  0.060 &  0.010 &  0.045 \\
Firm age (years)                  & -0.105 & -0.090 & -0.014 \\
\bottomrule
\end{tabular}%
}
\end{table}

\FloatBarrier

\begin{table}[t]
\centering
\caption{Top univariate predictors for each outcome based on signed AUC and signed PR-AUC computed on training rows. Signs indicate the direction of association with the outcome (positive or negative). Reported values reflect stand-alone discrimination, not additive effects. The results show that fundraising depends on recency and momentum, patent growth is mean-reverting with respect to age and IP stock, and exits are linked to accumulated financing maturity.}
\label{tab:predictors}
\resizebox{\textwidth}{!}{%
\begin{tabular}{lccc}
\toprule
{Outcome} & {Feature} & {AUC (signed)} & {PR-AUC (magnitude)} \\
\midrule
Next financing (12m) & Age (years) & 0.69 ($-$) & 0.31 \\
                     & Days since last round & 0.69 ($-$) & 0.27 \\
                     & Rounds in last 4q & 0.58 ($+$) & 0.23 \\
                     & Cumulative investors & 0.54 ($+$) & 0.22 \\
                     & Funding in last 4q (USD) & 0.57 ($+$) & 0.22 \\
\addlinespace
Patent growth (24m)  & Total patents & 0.70 ($-$) & 0.32 \\
                     & Total citations & 0.68 ($-$) & 0.31 \\
                     & Age (years) & 0.70 ($-$) & 0.39 \\
                     & Cumulative rounds & 0.66 ($-$) & 0.30 \\
                     & Days since last round & 0.61 ($-$) & 0.27 \\
\addlinespace
Exit (36m)           & Cumulative investors & 0.66 ($+$) & 0.12 \\
                     & Cumulative rounds & 0.64 ($+$) & 0.10 \\
                     & Cumulative capital raised (USD) & 0.65 ($+$) & 0.14 \\
                     & Days since last round & 0.61 ($-$) & 0.08 \\
                     & Late-stage rounds (cumulative) & 0.54 ($+$) & 0.08 \\
\bottomrule
\end{tabular}
}
\begin{flushleft}
\end{flushleft}
\end{table}
\FloatBarrier

\begin{table}[t]
\centering
\caption{Final feature sets used for model training by outcome, derived from the development period (2010–2019) after leakage-safe preprocessing and univariate screening. Each column lists the variables included in the persisted feature list used for downstream modeling and column alignment across splits. Features capture firm maturity, financing recency and momentum, and innovation intensity.}
\footnotesize
\resizebox{\textwidth}{!}{%
\begin{tabular}{p{0.31\textwidth} p{0.31\textwidth} p{0.31\textwidth}}
\toprule
{Next financing (12 months)} & {Patent growth (24 months)} & {Exit (36 months)}\\
\midrule
age & age & cumulative capital raised (USD) \\
days since last round & total patents & cumulative investors \\
rounds in last four quarters & total citations & cumulative rounds \\
cumulative investors & cumulative rounds & capital raised in last four quarters (USD) \\
capital raised in last four quarters (USD) & days since last round & cumulative late-stage rounds \\
cumulative capital raised (USD) & cumulative capital raised (USD) & days since last round \\
cumulative rounds & cumulative other-stage rounds & cumulative mid-stage rounds \\
total citations & cumulative investors & cumulative other-stage rounds \\
total patents & cumulative early-stage rounds & cumulative early-stage rounds \\
cumulative early-stage rounds & rounds in last four quarters & rounds in last four quarters \\
rounds this quarter & investors this quarter & total citations \\
investors this quarter & cumulative mid-stage rounds & total patents \\
capital raised this quarter (USD) & rounds this quarter & capital raised this quarter (USD) \\
cumulative mid-stage rounds & cumulative late-stage rounds & investors this quarter \\
cumulative other-stage rounds & capital raised this quarter (USD) & rounds this quarter \\
cumulative late-stage rounds & capital raised in last four quarters (USD) & age \\
\bottomrule
\end{tabular}
}
\label{tab:feature_subset}
\end{table}

\FloatBarrier

\begin{table}[t]
\centering
\footnotesize
\caption{Class imbalance and resampling feasibility on the development period (2010–2019). The table reports sample sizes, positive-class prevalence, and balanced sample counts under SMOTE-NC for each outcome. Only inverse-prevalence weighting and SMOTE-NC were carried forward into the main analysis; other resampling variants (ROS, Borderline-SMOTE, ADASYN) were retained for robustness checks but not used in final models. }
\label{tab:class_imbalance}
\resizebox{\textwidth}{!}{%
\begin{tabular}{lccccc}
\toprule
Outcome & Train $N$ & Positives & Prevalence & SMOTE\_NC $N$ (balanced) & Methods carried forward \\
\midrule
Next financing (12m)  & 2{,}495{,}444 & 439{,}009 & 0.176 & 4{,}112{,}870 & Weights; SMOTE-NC \\
Patent growth (24m)   & 2{,}495{,}444 & 547{,}553 & 0.219 & 3{,}895{,}782 & Weights; SMOTE-NC \\
Exit (36m)            & 2{,}495{,}444 & 147{,}395 & 0.059 & 4{,}696{,}098 & Weights; SMOTE-NC \\
\midrule
\multicolumn{6}{l}{\emph{Note:} Borderline-SMOTE and ADASYN error on \texttt{NaN}s and fall back to ROS; ROS saved but not used in main text.}\\
\multicolumn{6}{l}{Weights-only positive class weights (uncapped): 4.69 (12m), 3.56 (24m), 15.93 (36m); capped at 50 if needed.}\\
\bottomrule
\end{tabular}
}
\end{table}
\FloatBarrier

\begin{table}[t]
\centering
\caption{Winning model and out-of-time performance for each prediction task. Each outcome is evaluated on its corresponding non-overlapping split, with the positive-class base rate shown for reference. Reported metrics include AUROC and PR-AUC, with PR-AUC used as the main selection criterion. \emph{All reported winners use development-only preprocessing and inverse-prevalence weighting.}}

\label{tab:leaderboard}
\sisetup{round-mode=places,round-precision=3}
\resizebox{\textwidth}{!}{%
\begin{tabular}{l l c c c c c}
\toprule
{Outcome} & {Split (N)} & {Base rate} & {Model (imbalance)} & {AUROC} & {PR-AUC} & {Brier}  \\
\midrule
Next funding (12m) & Final 2022--2023 (281{,}326) & 0.088 & LGBM (weights) & 0.817 & 0.220 & 0.144  \\
Patent growth (24m) & Holdout 2020--2021 (644{,}765) & 0.139 & RF (weights) & 0.921 & 0.631 & 0.074  \\
Exit (36m) & Holdout 2020--2021 (273{,}147) & 0.048 & RF (weights) & 0.872 & 0.559 & 0.032  \\
\bottomrule
\end{tabular}
}
\vspace{0.5ex}
\end{table}

\FloatBarrier

\begin{table}[t]
  \centering
\caption{Precision at top-$K$ predictions (P@K) for each outcome on the corresponding evaluation split. Values reflect the proportion of true positives among the top-ranked predictions, compared to the base rate in the evaluated sample. Rankings are computed globally with deterministic tie-breaks. Precision may vary with $K$.}

  \label{tab:precision_at_k}
  \resizebox{\textwidth}{!}{%
  \begin{tabular}{l S[table-format=1.3] S[table-format=1.3] S[table-format=1.3] S[table-format=1.3]}
    \toprule
    {Outcome (split)} & {P@30} & {P@100} & {P@500} & {Base rate} \\
    \midrule
    Next funding (12m) (Final)  & 0.200 & 0.290 & 0.282 & 0.088 \\
    Patent (24m) (Holdout)      & 0.933 & 0.800 & 0.816 & 0.139 \\
    Exit (36m) (Holdout)        & 0.967 & 0.980 & 0.978 & 0.048 \\
    \bottomrule
  \end{tabular}
  }
  \vspace{0.25em}
\end{table}

\FloatBarrier

\begin{table}[t]
\centering
\footnotesize
\caption{Summary of final out-of-time scoring results for each prediction task. Each model was applied to the most recent evaluable cohort, using features and preprocessing fixed on the development period. Reported metrics correspond to the split used for model selection. Output files include organization identifiers, descriptors, predicted probabilities, ranks, and percentiles, with one record per firm after deduplication. Rankings can be used directly, while calibrated probabilities require post hoc isotonic scaling. \emph{Selection metrics are repeated from Table~\ref{tab:leaderboard}; no re-estimation is performed. For exits, the evaluable holdout rows are concentrated in 2020 due to the 36-month horizon.}}

\label{tab:scoring_outputs}
\resizebox{\textwidth}{!}{%
\begin{tabular}{l l l c c}
\toprule
Outcome & Winner (imbalance; model) & Selection split and metrics & Scored cohort & Rows \\
\midrule
Next funding (12 months) & LGBM (weights) & Final: PR-AUC 0.220, AUROC 0.817 & 2022 & 281{,}326 \\
Patent growth (24 months) & RF (weights) & Holdout: PR-AUC 0.631, AUROC 0.921 & 2021 & 280{,}249 \\
Exit (36 months) & RF (weights) & Holdout: PR-AUC 0.559, AUROC 0.872 & 2020 & 273{,}147 \\
\bottomrule
\end{tabular}
}
\vspace{0.5em}
\footnotesize
\end{table}
\FloatBarrier

\end{document}